\documentclass[10pt,twocolumn,letterpaper]{article}

\usepackage{iccv}
\usepackage{times}
\usepackage{epsfig}
\usepackage{graphicx}
\usepackage{amsmath}
\usepackage{amssymb}
\usepackage[symbol*]{footmisc}
\usepackage{color}
\usepackage{colortbl}
\definecolor{myred}{RGB}{255,222,222}

\usepackage[pagebackref=true,breaklinks=true,letterpaper=true,colorlinks,bookmarks=false]{hyperref}

\iccvfinalcopy 

\newif\ifrebuttal
\rebuttaltrue

\def\iccvPaperID{1759} 

\newcommand{\D}{{\cal D}}

\newcommand{\N}{{\cal N}}
\newcommand{\C}{{\cal C}}

\newcommand{\y}{{\bf y}}

\makeatletter
\usepackage{xspace}
\def\@onedot{\ifx\@let@token.\else.\null\fi\xspace}
\DeclareRobustCommand\onedot{\futurelet\@let@token\@onedot}

\newcommand{\figref}[1]{Fig\onedot~\ref{#1}}

\newcommand{\tabref}[1]{Tab\onedot~\ref{#1}}

\def\eg{\emph{e.g}\onedot} 
\def\ie{\emph{i.e}\onedot} 
 
\def\etc{\emph{etc}\onedot} 
 
\def\etal{\emph{et al}\onedot}

\makeatother

\DeclareMathOperator*{\argmax}{\arg\!\max}

\begin{document}

\title{Monocular Object Instance Segmentation and Depth Ordering with CNNs}

\author{Ziyu Zhang\thanks{The first two authors contributed equally to this work.} \hspace{0.4cm} Alexander~G. Schwing\footnotemark[1] \hspace{0.4cm} Sanja Fidler \hspace{0.4cm} Raquel Urtasun \\
Department of Computer Science, University of Toronto \\
{\tt\small \{zzhang, aschwing, fidler, urtasun\}@cs.toronto.edu}
}

\maketitle
\thispagestyle{empty}

\begin{abstract}
In this paper we tackle the problem of instance-level segmentation and depth ordering from a single monocular image. Towards this goal, we take advantage of convolutional neural nets and train them to directly predict instance-level segmentations where the instance ID encodes the depth ordering within image patches. 
To provide a coherent single explanation of an image we develop a Markov random field which takes as input the predictions of convolutional neural nets applied at overlapping patches of different resolutions, as well as the output of a connected component algorithm. It aims to predict accurate instance-level segmentation and depth ordering. We demonstrate the effectiveness of our approach on the challenging KITTI benchmark and show good performance on both tasks. 
\end{abstract}

\section{Introduction}

Over the past few decades, two main tasks for parsing visual scenes have received a lot of attention: 
object detection where the goal is to place bounding boxes accurately around each object, and pixel-level labeling which aims to assign a class label to each pixel in the image. We follow some of the recent work~\cite{HariharanSimultaneousECCV2014,yang2012layered,tigheCVPR14,HeCVPR2014,WangCVPR15}, however we argue that the next generation of recognition techniques should provide a more detailed parsing of a scene by labeling each object instance in an image with an accurate segmentation, a generic class, and 3D information such as depth ordering. This is particularly important for applications such as driver assistance, where an ideal system needs to be aware of each individual object and their spatial arrangements. It is also important for, \eg, image captioning, Q\&A and retrieval techniques~\cite{Karpathy14,malinowski14nips,LinCVPR14}, where describing a 3D scene is easier and potentially more informative than describing a soup of orderless object detections~\cite{Karpathy14}.

The goal of this paper is to predict an accurate pixel-level labeling of each object instance belonging to the class of interest, as well as their depth ordering, given  a single monocular image.  
This is a very challenging problem since it requires us to jointly solve for the class labeling of each pixel, and  their combinatorial grouping  into objects. Moreover, reasoning about depth from a single image is known to be a difficult and ill-posed problem, where object-level priors are typically needed to resolve ambiguities~\cite{in-perspective}.

\begin{figure}[t]
\centering
\includegraphics[width=8cm,trim = 1mm 20mm 2mm 18mm, clip=true]{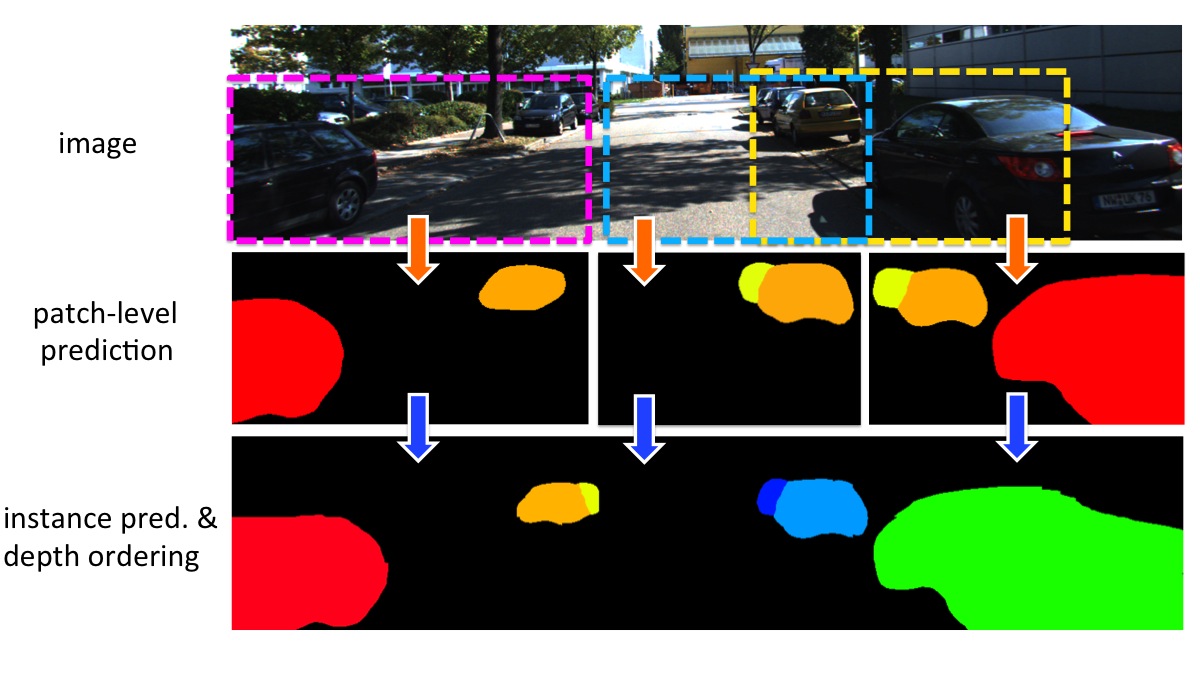}
\caption{Our approach uses a CNN to predict instance-level segmentation and depth ordering in an image patch. We sample a stride of patches at several scales, and combine predictions into the final labeling by solving an energy minimization problem.}
\label{fig:intro}
\end{figure}

In this paper, we tackle the  autonomous driving scenario~\cite{kitti} and focus on cars which are particularly challenging in this domain. Each image typically contains many car instances forming complex occlusion patterns. Shadows, reflectance and saturation present an additional challenge. 
We formulate the problem as the one of inference in a Markov Random Field (MRF) that reasons  jointly about  pixel-wise instance level segmentation and depth ordering. We build on the success of convolutional neural nets (CNNs) to object segmentation~\cite{LongCVPR2014,GuistiICIP2013,SermanetICLR2014,JayICLR2015,GeorgeTR2015,SchwingTR2015,ZhengARXIV2015} and define our unary and pairwise potentials using the output of convolutional neural networks at multiple resolutions: our network operates on densely sampled image patches and is trained to predict a depth-ordered instance labeling of the patch. We refer the reader to \figref{fig:intro} for an illustration. We sample a stride of patches at several scales, and combine predictions into the final labeling via the MRF. 
Our energy terms encode the fact that connected components detected within a patch should be ordered, and the affinity of neighboring pixels depends on the CNN. 
In contrast to the majority of existing work~\cite{tigheCVPR14,yang2012layered,HariharanSimultaneousECCV2014}, no object detections are needed as input to our method as we reason about detection and segmentation jointly. 
Our approach uses 3D information (3D bounding boxes and stereo) during training, but requires only a single RGB image at test time.
We exploit~\cite{ChenCVPR14} to infer object segmentations from available 3D bounding box annotations which allows us to train our network on a  large scale dataset.

We evaluate our method on the subset of the KITTI detection benchmark~\cite{kitti}, labeled with car instance segmentations~\cite{ChenCVPR14}. 
Our experimental evaluations  show that our network is capable of accurate prediction for up to 5 object instances in an image patch. 
In the remainder of the paper, we review related work, explain our CNN architecture, present our approach and detail the obtained results. 


\begin{figure}[t]
\centering
\includegraphics[width=\linewidth]{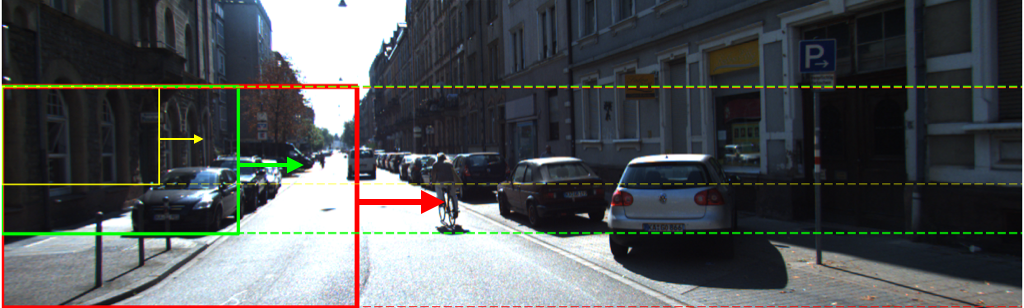}
\caption{The patches extracted from an image at a large (red), medium (green) and small (yellow) scale.}
\label{fig:PatchVisualization}
\end{figure}


\section{Related Work}

We focus our review on techniques operating on a single monocular image. Our contribution is related to a  line of work that aims at predicting depth ordering  using weak object information as well as mid-level cues such as junctions and boundaries~\cite{HoiemICCV07,JiaCVPR12,KowdleCVPR13}. These approaches typically do not reason about class and instance segmentation, but aim at assigning each pixel to a particular depth layer. 

Combination of detection and segmentation has been the most common approach to instance segmentation: once an object is localized with a bounding box, top-down information such as shape or appearance can be used to carve out the object within the box~\cite{Kumar05}. In~\cite{ladicky10,whole}, the authors propose an energy minimization framework which jointly identifies the true positive boxes and labels the pixels with the class labels.~\cite{HaykoECCV12} uses Hough voting to obtain object centers and proposes a CRF model  to jointly reason about the assignment of pixels to centers (which define an object instance) and predict its semantic class. 

Approaches that infer 3D objects from RGB images inherently reason about instance detection and depth ordering, but typically not segmentation. Hoiem~\etal~\cite{in-perspective} and Wang~\etal~\cite{WangCVPR15} infer 3D objects from 2D detections using several geometric cues (ground plane priors, horizon, typical physical sizes, \etc). In~\cite{xiangcvpr15},  2D boxes are detected and the method reasons  about learned 3D occlusion patterns within each box. Occlusion-aware methods implicitly reason about instances and their depth ordering. A popular approach has also been to use a collection of CAD models and match them to RGB images~\cite{Aubry14,LimICCV13}. Online object catalogs can be used to assign physical size to the CAD models based on their semantic class, giving rise to a 3D interpretation of the scene.  In indoor scenarios, the Manhattan world assumption has been leveraged frequently to reason about the 3D room and object cuboids~\cite{Choi13, box,lee,HedauCVPR12}. This yields instance segmentation as well as depth ordering, however, segmentation is typically limited to a simple cuboid-like object shape.

Recently several approaches emerged to tackle the problem of object instance segmentation. The powerful R-CNN framework~\cite{girshick2013rich} was employed by~\cite{HariharanSimultaneousECCV2014} to detect objects.  Two networks were used to segment the objects contained in the boxes. \cite{SilbermanECCV14} tries to make an optimal cut in the hierarchical segmentation tree to yield object instance regions. 

Most related to our work are~\cite{tigheCVPR14,Isola13,yang2012layered} which explicitly reason about class and instance segmentations as well as depth ordering. Tighe~\etal~\cite{tigheCVPR14} first  infer a pixelwise class labeling of the image and  detect objects. Then they solve for instances and depth ordering by minimizing an integer quadratic program. In contrast, our approach directly predicts  instance labeling and depth ordering via a powerful CNN network, and a merging procedure phrased as an energy minimization task. 
A ``scene collage'' model is proposed in~\cite{Isola13}, that, for a given test image, retrieves the closest visualization from a dataset and aims to transform its object masks into a proper 3D scene interpretation. In~\cite{yang2012layered}, object detectors first provide candidate locations and object masks, and a probabilistic model is proposed to order them coherently in depth layers. 

\section{Instance Segmentation and Depth Ordering}


Our goal is to predict an accurate pixel-level labeling of each object instance  given a  single monocular image. In addition, we want to order the objects of interest according to their distance from the camera. 
We formulate this task jointly as a pixel-level labeling where each state denotes an instance and its label ID encodes the ordering. 

Toward this goal, we take advantage of deep learning  and train a CNN to predict both  instance segmentations and  depth ordering.
To deal with the different scales of objects, we split a given image into a set of overlapping patches, extracted at multiple scales, and  employ the CNN to estimate a pixel-level labeling and the depth ordering for each patch. This provides a distribution across different depth levels as well as  the background state  for each patch. We visualize the extraction of the patches in \figref{fig:PatchVisualization}. In particular, we horizontally tile the  image below a pre-specified height  into the largest patches (red rectangle). For medium (green) and small (yellow) size patches we only tile the region around the pre-specified height, as  we expect to see small cars only close to the horizon. This is due to the fact that the KITTI dataset~\cite{kitti} was acquired with the camera being mounted on top of a car and thus the vertical axis in the image is correlated with depth.  

Training deep networks typically requires large datasets. To our knowledge there is unfortunately no such dataset publicly available to date. Therefore we take advantage of the weakly labeled approach of~\cite{ChenCVPR14} which takes as input a 3D bounding box, LIDAR points and stereo imagery, and generates car segmentations with accuracy as high as mechanical turk. This allows us to take advantage of the KITTI dataset, from which we use 6,446 training images containing 24,724 cars. 

Given the predictions of the network for different overlapping patches, we need to create a single coherent labeling for the image. Towards this goal, we formulate the problem as inference in an MRF, encoding the segmentation of objects and the ordering of depth. Importantly we emphasize that our MRF is not specifically tailored to the way patches are extracted, \ie, we can fuse an arbitrary patch configuration. 
In the remainder of the section, we first introduce our CNN and then discuss our energy minimization framework.

\subsection{Instance Segmentation and Depth Ordering Network}

Let  $x$ be the input image. 
We are interested in predicting for every pixel a label indicating the instance it belongs to. 
Thus for the $p$-th pixel, we represent its state with $y_p \in \{0, \ldots, N\}$, with $N$ the maximum number of cars that can be present in an image. 
Furthermore, for each image, the instances are ordered in depth, \ie, an instance assigned state $i$ is closer to the camera than an object  assigned state $j$ if and only if  $i<j$.



Given mean-subtracted, differently sized and overlapping patches extracted from an image, we perform for each patch a forward pass through a CNN. The output of the network is a score $F(z,\y_z,w)$ which depends on the input patch $z$, the parameters $w\in\mathbb{R}^A$ of the network and the considered pixel-wise depth-level map $\y_z$, restricted to the pixels in the patch $z$. 

\begin{figure}[t]
\centering
\includegraphics[width=0.625\linewidth]{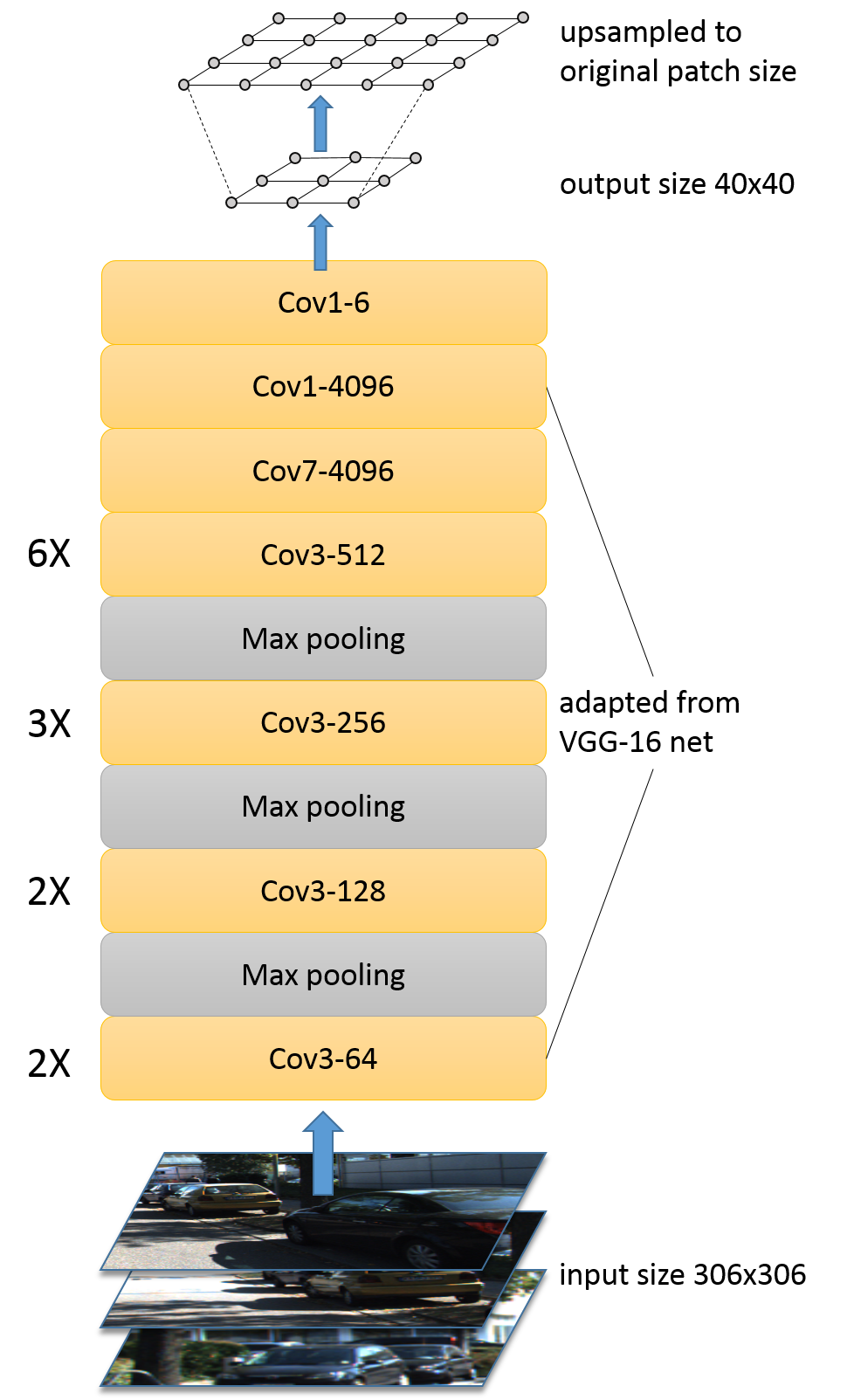}
\caption{Our convolutional neural network.}
\label{fig:NetworkVisualization}
\vspace{-2mm}
\end{figure}

\begin{figure*}[t]
\centering
\vspace{-2mm}
\begin{minipage}{0.73\linewidth}
\includegraphics[height=3.0cm]{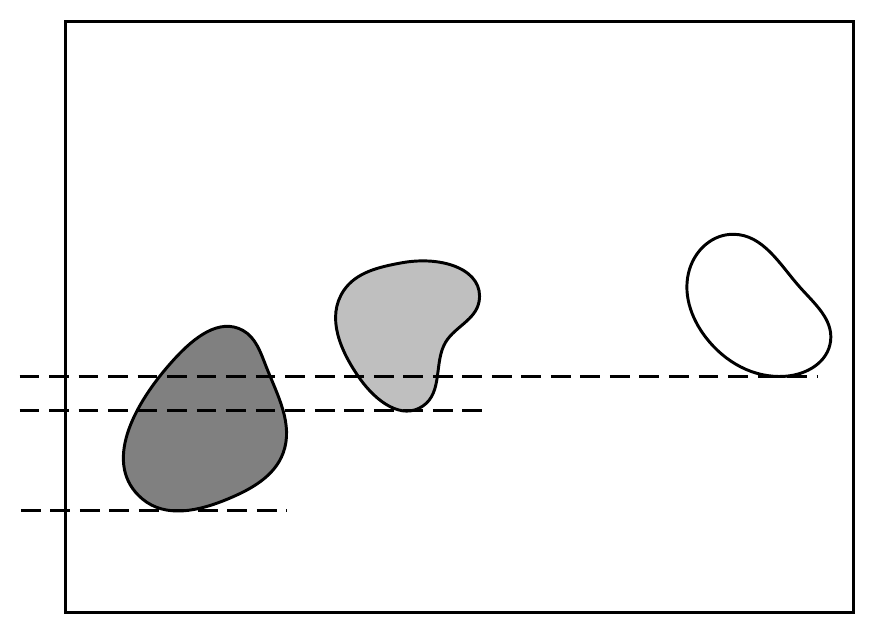}
\includegraphics[height=3.0cm]{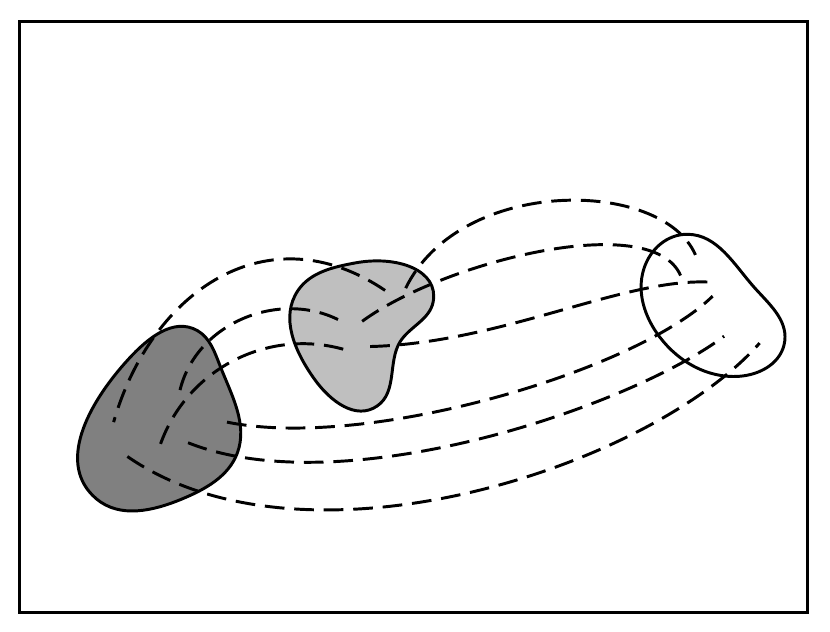}
\includegraphics[height=3.0cm]{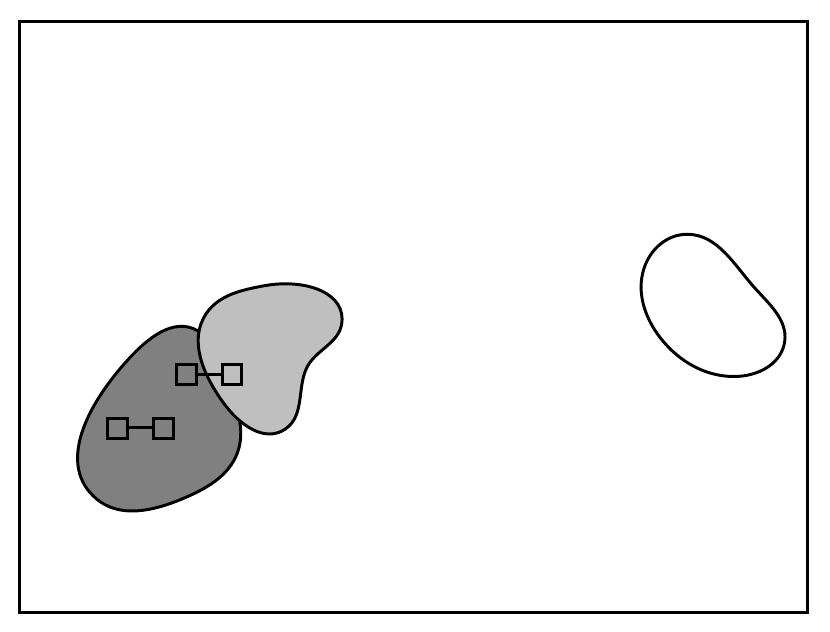}
\end{minipage}
\begin{minipage}{0.26\linewidth}
\caption{Illustration of intuition behind the unary conn. component energy (left), long-range (middle) and short-range connections (right). 
}
\label{fig:PotentialIllustration}
\end{minipage}
\end{figure*}

To design such a network we start from the 16-layer VGG network~\cite{SimonyanARXIV2014}, which was originally designed to predict a single categorical variable for a given input patch of size $224\times 224$. In contrast, we need a pixel-wise prediction, \ie, a categorical variable for each pixel. Therefore we first convert the three fully-connected layers of the VGG network into convolutional units as in~\cite{GuistiICIP2013,SermanetICLR2014,LongCVPR2014}. After this conversion, an increase of the size of the input patches to $306\times 306$ translates into an increased output-space, \ie, we obtain multiple categorical variables. Importantly, computational efficiency is retained even after this transformation.  However, due to the $5$ pooling layers, each downsampling its input by a factor of two, we end up with a rather coarse output being $2^5$ times smaller than the input patch size. 

To compensate for this loss in resolution,  we follow~\cite{JayICLR2015} and skip downsampling at the two top-most pooling layers, to obtain an output being only $8$ times smaller than the input. To ensure that the computations remain identical we note that pooling at a stride of one rather than two effectively produces four images having their pixels stored in a spatially interleaved manner. Therefore, the subsequent convolutional layers need to take into account the fact that data is no longer consecutively stored in memory. Data subsampling can be efficiently implemented for convolutions, which has been previously discussed in the work by Mallat~\cite{Mallat1999}, known as the `\`{a} trous (with hole) algorithm' and applied to CNNs in recent work~\cite{JayICLR2015,GeorgeTR2015,SchwingTR2015}. We visualize the resulting network in \figref{fig:NetworkVisualization}.

We assume a maximum of 6  instance levels (including background) to be present in any given patch. Therefore we also replace the top-layer matrix of the VGG network with a corresponding convolution producing an output map of size $40 \times 40 \times 6$.


To find good parameters of our network for the depth-ordering task, we take advantage of a training set $\D = \{(z,\y^\text{\tiny{GT}}_z)\}$ containing $|\D|$ pairs of patches $z$, \ie, extracted pixels of enlarged, car containing ground-truth bounding boxes, and corresponding ground-truth depth-ordering maps $\y^\text{\tiny{GT}}_z$. Importantly, we extract patches and corresponding depth-level maps at different resolutions. 
To benefit from efficient GPU computation (\ie, consistent mini-batches) we re-scale all patches within the dataset $\D$ to a size of $306\times 306$ pixels and downsample the ground-truth to $40\times 40$ pixels. We learn the parameters $w$ of the CNN  by optimizing cross-entropy.
The resulting program is non-convex and we follow existing literature by using a simple yet effective stochastic gradient descent procedure. Because of the GPU memory requirements we train with a batch-size of $5$ using a weight-decay of $0.0005$, a moment of $0.9$. The initial stepsize is $0.01$ for the top layer and $0.001$ for all subsequent parameters.



\subsection{MRF Patch Merging}

We next need to combine the output of the CNN at different overlapping patches to produce a single coherent explanation of the image. We formulate the problem as the one of inference in a Markov random field which reasons about multiple object instances and their ordering in depth. 

Let $y_p\in \{0, \ldots, N\}$ denote the depth level prediction of pixel $p$ in a given image. 
We use a maximum of $N=9$ cars per image. 
Let us emphasize that the CNN produces a prediction of up to 5 cars for a patch in the image and the goal of the MRF is to create a single coherent  explanation. 
Let $\y$ be the joint labeling for each pixel in the image. 

To estimate the high level structure, we first run a connected component algorithm on the merged CNN prediction map. Merging is performed by first averaging for each patch the predicted probabilities within each connected component of the patch prediction. Afterwards we compute the merged map for each pixel by finding the maximizing state across possible pixel labels and patches. 

We then define the MRF energy to encode the fact that the final labeling should be consistent with the local CNN predictions and the connected components in terms of both the instances as well as the depth ordering. We thus define
\small
\begin{eqnarray*}
E(\y) &=& \sum_p \left(E_{\text{CNN},p}(y_p) + E_{\text{CCO},p}(y_p)\right) \\
&+& \sum_{p,p' : \C(p)\neq\C(p')} E_{\text{long},p,p'}(y_p,y_{p'}) \\
&+& \sum_{p,p'\in \N(p)} E_{\text{short},p,p'}(y_p,y_{p'}),
\end{eqnarray*}
\normalsize
with $\N(p)$ a 4-neighborhood and $\C(p)$ the connected component that pixel $p$ belongs to. 
It is important to note that the patch predictions that we obtain from the CNN are downsampled by a factor of $8$. In order to correct for this reduced resolution,  we bi-linearly interpolate each CNN patch output back to its original size.  
We now explain the different terms in more detail. 

\begin{table*}[t]
\vspace{-0.25cm}
\centering
\small

    \begin{tabular}{| c | c | c | c | c | c | c |}
    \hline
    Exp & FIoU & BIoU & AvgIoU & Acc & OvrlPr & OvrlRe \\
    \hline\hline
    large & 75.2 & 97.9 & 86.5 & 98.0 & 85.7 & 85.9 \\
    \hline
    medium & 80.0 & 98.0 & 89.0 & 98.1 & 87.6 & 90.2 \\
    \hline
    small & 75.0 & 98.0 & 86.5 & 98.1 & 85.7 & 85.7 \\
    
    \hline\hline
    \cite{ohn2015learning} & 38.3 & 96.6 & 67.4 & 96.7 & 88.0 & 40.4\\
    \hline\hline
    CNNRaw & 80.6 & {\bf 98.8} & 89.7 & {\bf 98.9} & 88.0 & 90.6\\
    \hline
    Unary & 80.6 & {\bf 98.8} & 89.7 & {\bf 98.9} & 88.0 & 90.6\\
    \hline
    Unary+ShortRange & {\bf 80.8} & {\bf 98.8} & {\bf 89.8} & {\bf 98.9} & 87.4 & 91.4\\
    \hline
    Full & 80.7 & {\bf 98.8} & {\bf 89.8} & {\bf 98.9} & 86.6 & {\bf 92.2}\\
    \hline\hline
    CNNRaw+PP & 80.4 & {\bf 98.8} & 89.6 & {\bf 98.9} & {\bf 88.2} & 90.2\\
    \hline
    Unary+PP & 80.6 & {\bf 98.8} & 89.7 & {\bf 98.9} & 88.1 & 90.4\\
    \hline
    Unary+ShortRange+PP & 80.6 & {\bf 98.8} & 89.7 & {\bf 98.9} & 87.5 & 91.1\\
    \hline
    Full+PP & 80.5 & {\bf 98.8} & 89.7 & {\bf 98.9} & 87.0 & 91.5\\
    \hline
    
%
    
    \end{tabular}
    \caption{Class-level segmentation results on the test set.}
		\label{tab:binaryMeasuresTest}
		\vspace{-0.4cm}
\end{table*}

\vspace{-0.3cm}
\paragraph{CNN Energy: } This term encodes the fact that the local CNN prediction is always lower than the true global one. This is valid since the CNN has only access to a subset of the image, illustrating a subset of cars. 
We thus include a unary potential per pixel that favors all states equal to or higher than the one predicted by the CNN in each local patch.
Formally, we define the unary energy as a sum over patches, $E_{\text{CNN},p}(y_p) = \sum_z E_{\text{CNN},z,p}(y_p)$, with
\small
\begin{equation*}
E_{\text{CNN},z,p}(y_p) = \begin{cases}
-1& \text{if } y_p \geq y^*_{z,p} \\
0 & \text{otherwise}
\end{cases},
\end{equation*}
\normalsize
where $\y^*_z = \argmax\limits_{\y_z} F(z,\y_z,w)$.

\vspace{-0.3cm}
\paragraph{Connected Components Ordering:} 
We order the connected components according to the vertical axis. For traffic scenarios this is an indication of depth. For each pixel, this term favors states that are equal to or larger than the order of its assigned connected component. Note that we do not include the background as a component. 
This is illustrated in \figref{fig:PotentialIllustration} (left) for an example with 3 connected components (\ie.,  3 cars).
Our potential is formally defined as:
\small
\begin{equation*}
E_{\text{CCO},p}(y_p) = \begin{cases}
-1 & \text{if } y_p \geq O(p) \\
0 & \text{otherwise}\end{cases},
\end{equation*}
\normalsize
where $O(p)$ denotes the order of the connected component assigned to pixel $p$.

\vspace{-0.3cm}
\paragraph{Long-range Connections}
This term prefers pixels to be assigned to different states if they belong to different components. Since fully connecting all the pixels within pairs of components would make inference very slow, we randomly sample 20,000 connections  per  component pair. 
This is illustrated in \figref{fig:PotentialIllustration} (center) for an example with 3 connected components (\ie,  3 cars) and 3 sampled connections.
The potential is formally given via
\small  
\begin{equation*}
\begin{split}
& E_{\text{long},p,p'}(y_p,y_{p'})\\ 
& = \begin{cases}
-1& \text{if } y_{p'}>y_p, y_p \neq 0, O(p')>O(p) \\
0 & \text{otherwise}
\end{cases}.
\end{split}
\end{equation*}
\normalsize

\vspace{-0.3cm}
\paragraph{Short-range Connections:} This term encourages nearby pixels to be assigned the same labeling if this is also the case in the CNN predictions. On the other hand, if the CNN predicts different states for two  nearby pixels, we encourage them to be different. 
This can be done via a weighted Potts-type potential using the connectivity illustrated in \figref{fig:PotentialIllustration} (right), where we highlight neighboring pixels using two rectangles connected by a line.

\subsection{MRF Inference}

Inference in our MRF is NP-hard, since we formulated a multi-label problem with attractive and repulsive potentials. 
Inspired by the $\alpha$-$\beta$-swap algorithm, we designed an inference procedure that iterates between changes of two sets of labels. This subproblem is non-submodular in our case, as our energy is not regular. As a consequence we resort to quadratic pseudo-boolean optimization (QPBO) to solve the binary inference problem. We utilize a default value for the nodes that QPBO is not able to label. Importantly, we accept the move only if the energy decreases. 


%


\subsection{Post processing}
Given the result of the previous section, we perform a few post-processing steps to improve the results. First we remove spurious isolated groups of object instances smaller than 200 pixels. This step is based on the intuition that objects of interest are of a certain size.
We then perform hole-filling, \ie, for each object instance containing a hole, we simply fill in the hole using the label of the surrounding object.
Finally, we re-label disconnected instance labelings and re-order them according to the vertical axis coordinate of the center of the 2D bounding box around each connected component.

\begin{table*}[t]
\vspace{-0.3cm}
\centering
\small
    \begin{tabular}{| c | c | c | c | c | c | c | c | c |}
    \hline
    Exp & MWCov & MUCov & AvgPr & AvgRe & AvgFP & AvgFN & ObjPr & ObjRe \\
    \hline\hline
    large & 36.7 & 29.3 & 81.1 & 53.0 & 0.0751 & 0.457 & 49.5 & 36.4 \\
    \hline
    medium & 40.3 & 35.2 & 78.1 & 68.1 & 0.115 & 0.174 & 57.7 & 52.6 \\
    \hline
    small & 34.6 & 31.1 & 77.0 & 67.1 & 0.124 & 0.126 & 56.4 & 55.8 \\
    
    
    \hline\hline
    \cite{ohn2015learning} & 45.4 & 40.1 & {\bf 86.3} & 45.9 & {\bf 0.015} & 1.650 & {\bf 95.1} & 48.5\\
    \hline\hline
    CNNRaw & 51.1 & 33.3 & 82.5 & 65.3 & 0.081 & 0.569 & 42.9 & 21.9\\
    \hline
    Unary & 69.2 & 53.8 & 82.8 & 65.3 & 0.198 & 0.569 & 68.3 & 58.1\\
    \hline
    Unary+ShortRange & 54.5 & 37.2 & 81.8 & 66.8 & 0.117 & 0.533 & 47.4 & 26.1\\
    \hline
    Full & 68.1 & 53.5 & 81.7 & {\bf 68.8} & 0.168 & {\bf 0.528} & 64.7 & 56.1\\
    \hline\hline
    CNNRaw+PP & 68.8 & 53.4 & 83.4 & 63.5 & 0.274 & 0.751 & 60.1 & 56.2\\
    \hline
    Unary+PP & 69.6 & 54.1 & 83.2 & 64.3 & 0.259 & 0.751 & 66.6 & 58.3\\
    \hline
    Unary+ShortRange+PP & {\bf 71.3} & {\bf 55.9} & 82.5 & 65.5 & 0.320 & 0.685 & 63.0 & 58.8\\
    \hline
    Full+PP & 70.3 & 55.4 & 79.2 & 66.5 & 0.411 & 0.675 & 59.2 & {\bf 59.0}\\
    \hline
    
    
    
    \end{tabular}
    \caption{Instance-level metrics on the test set.}
		\label{tab:InstanceMeasuresTest}
		\vspace{-0.2cm}
\end{table*}


\begin{table*}[t]
\centering
\small
    \begin{tabular}{| c | c | c | c | c | c | c | c |}
    \hline
    Exp & \#Ins & \%RcldIns & \#InsPair & \%RcldInsPair & InsPairAcc & \%CorrPxlPairFgr  \\


    \hline\hline
    large & 1749 & 36.4 & 2085 & 9.4 & 99.0 & 80.8 \\ 
    \hline
    medium & 1977 & 52.6 & 1662 & 17.1 & 98.2 & 82.7  \\ 
    \hline
    small & 2176 & 55.8 & 1318 & 21.1 & 96.8 & 74.2 \\ 
    \hline\hline
    \cite{ohn2015learning}+Y & 804 & 48.5 & 1740 & 21.1 & 92.4 & 14.8\\
    \hline
    \cite{ohn2015learning}+Depth & 804 & 48.5 & 1740 & 21.1 & 94.3 & 14.8\\
    \hline
    \cite{ohn2015learning}+Size & 804 & 48.5 & 1740 & 21.1 & 89.4 & 14.8\\
    \hline\hline
    CNNRaw & 804 & 21.9 & 1740 & 1.7 & 100.0 & 57.5\\
    \hline
    Unary & 804 & 58.1 & 1740 & 28.2 & 92.4 & 80.9\\
    \hline
    Unary+ShortRange & 804 & 26.1 & 1740 & 3.4 & 100.0 & 62.1\\
    \hline
    Full & 804 & 56.1 & 1740 & 25.5 & 93.0 & 77.4\\
    \hline\hline
    CNNRaw+PP & 804 & 56.2 & 1740 & 26.6 & 83.2 & 79.0\\
    \hline
    Unary+PP & 804 & 58.3 & 1740 & 28.3 & 92.5 & 81.2\\
    \hline
    Unary+ShortRange+PP & 804 & 58.8 & 1740 & {\bf 29.5} & 84.8 & 81.4\\
    \hline
    Full+PP & 804 & {\bf 59.0} & 1740 & 29.3 & 90.4 & {\bf 83.1}\\ 
    \hline
    
    
    
    
    \end{tabular}
    \caption{Depth ordering assessment on the test set.}
		\label{tab:DepthMeasuresTest}
		\vspace{-0.4cm}
\end{table*}

\section{Experimental Evaluation}

We used the challenging KITTI benchmark~\cite{kitti} for our experiments. 
To evaluate all the approaches we employ the car segmentation ground-truth of~\cite{ChenCVPR14}, which consists of $301$ images labeled by in-house annotators, providing very high-quality pixel-wise labeling for a total of $1,229$ cars. 
The rest of the $6,744$ images from the training set of KITTI's object detection benchmark are employed for training the CNN.



\vspace{-0.4cm}
\paragraph{Training the CNN: } Training a multi-layer convolutional network requires a sufficiently large training set. We thus derive a high-quality surrogate ground truth using the method of~\cite{ChenCVPR14}. It solves a submodular energy that makes use of ground-truth 3D bounding box annotations, point cloud, stereo imagery and shape priors from rendered CAD models~\cite{cadmodel} in order to generate a pixel-wise labeling of the object. To train the weights that combine different potentials in~\cite{ChenCVPR14} we utilize 3 annotated images. 
We then apply the method to provide the labeling of all $6,744$ images minus the $301$ samples that have accurate segmentations, and train the CNN on this subset augmented by the three annotated images used for learning the model of~\cite{ChenCVPR14}. We compute depth ordering by sorting the 3D bounding boxes that are provided in  KITTI. Since we know the distance of each 3D bounding box to the camera, we can order the car instances accordingly. 

The remaining $298$ out of $301$ annotated images are divided into $101$ validation and $197$ test images, and we ensure that the validation set and the test set have  roughly the same distribution of the total number of  car instances within an image. Importantly, all our hyper-parameters are tuned on the validation set only, \ie, we only evaluated our model on the held-out test set once to generate the numbers reported below. 
We provide results for the three patch scales large, medium and small evaluated on a patch-based level. We also provide an ablation study on our MRF approach both with (denoted by `+PP') and without post-processing, and compare to the instance-level segmentation approach derived from~\cite{ohn2015learning} which, taken their detection and orientation estimation, generates a 3D bounding box and projects a fitted CAD model back to the image space\footnote[2]{We didn't succeed in running the method of~\cite{tigheCVPR14}.
}. 
In the ablation analysis we provide the performance of the CNN output (CNNRaw) with different MRF formulations containing only unary (Unary), containing only unary and short range (Unary+ShortRange) or containing all defined energy terms (Full).


We assess the quality of our approach using a wide range of  metrics which we describe briefly in the following. Since we combine pixel-wise instance-level segmentation with depth ordering we need to compare using metrics that assess the respective performance. We divide our evaluation into three parts, `class-level' measures, `instance-level' metrics and `depth ordering' assessment.

\vspace{-0.4cm}
\paragraph{Class-level segmentation:}  We first assess the foreground-background prediction performance of our approach, \ie, how accurately we are able to differentiate cars from other objects and the background. We use the intersection over union (IoU) metric and provide foreground IoU (FIoU) evaluating the accuracy of the car detections, background IoU (BIoU) assessing the accuracy of the background prediction, the average of both referred to as average IoU (AvgIoU). We also provide pixel-wise prediction accuracy (Acc), overall precision (OvrlPr) being the number of true positive pixels over the sum of true and false positive pixels, and overall recall (OvrlRe) being the number of true positive pixels over the sum of true positive and false negative pixels.
We summarize our results on the test set in \tabref{tab:binaryMeasuresTest}.  
We note that the medium sized patches seem to perform slighly better on average than the small and large ones.
We also emphasize that the three top rows, evaluated on patch level, are not directly comparable to the bottom nine rows, \ie, only the bottom nine rows are evaluated on the image level.
We observe that our full MRF formulation achieves the best average IoU irrespective of whether we apply post-processing or not and performs well on other IoU metrics, accuracy as well as recall. The approach derived from~\cite{ohn2015learning} achieves a better precision but has much worse recall. We also observe the post-processing to harm the binary prediction performance in general.

\begin{figure*}[t]
\vspace{-0.4cm}
\newlength{\MyImgWidth}
\setlength{\MyImgWidth}{5cm}
\centering


\includegraphics[width=\MyImgWidth]{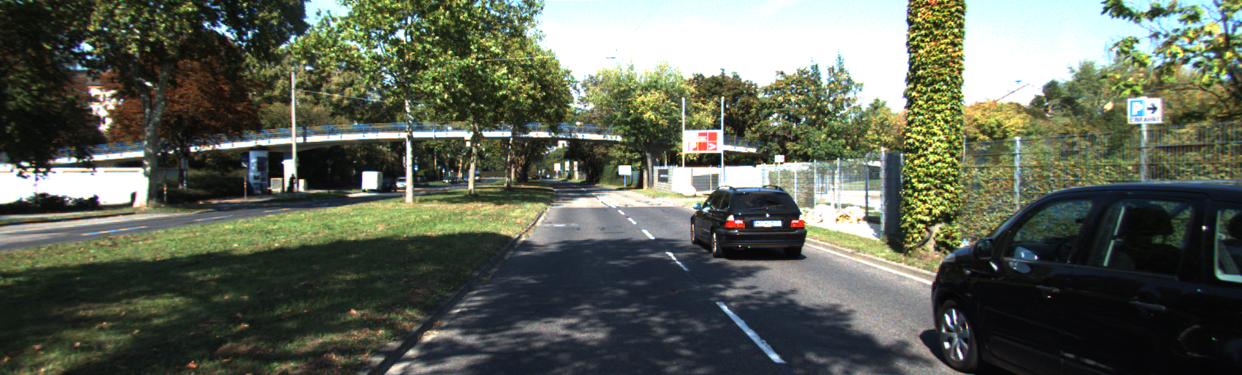}
\includegraphics[width=\MyImgWidth]{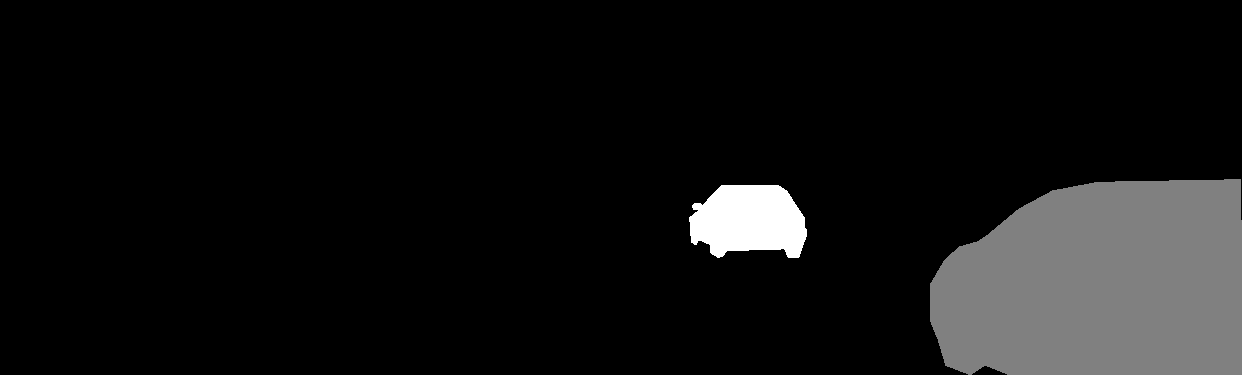}
\includegraphics[width=\MyImgWidth]{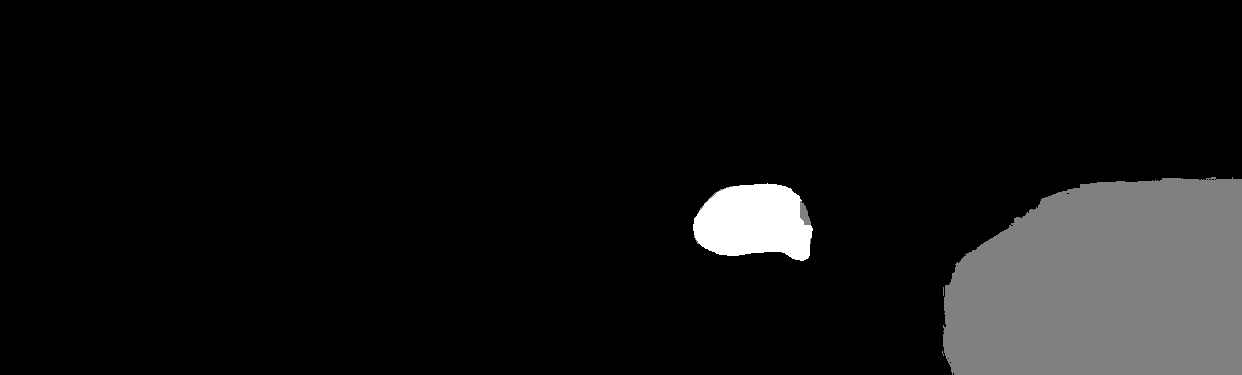}

\includegraphics[width=\MyImgWidth]{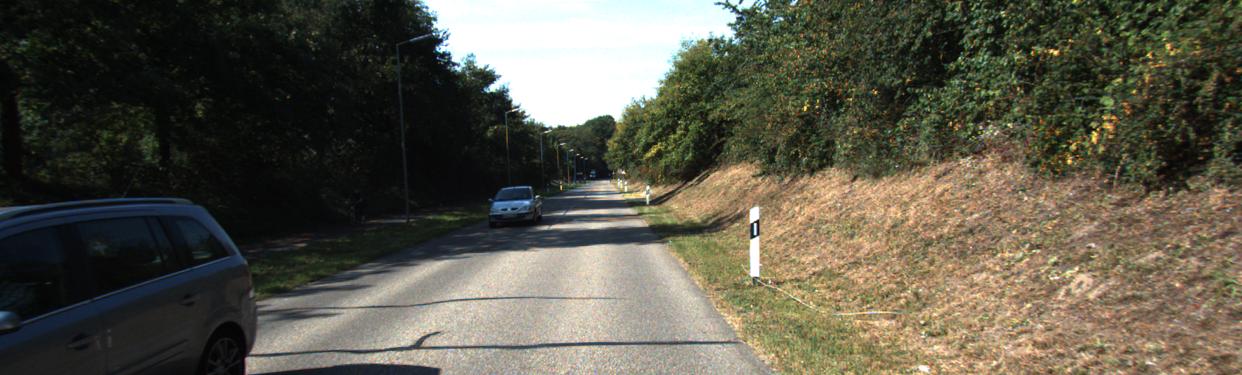}
\includegraphics[width=\MyImgWidth]{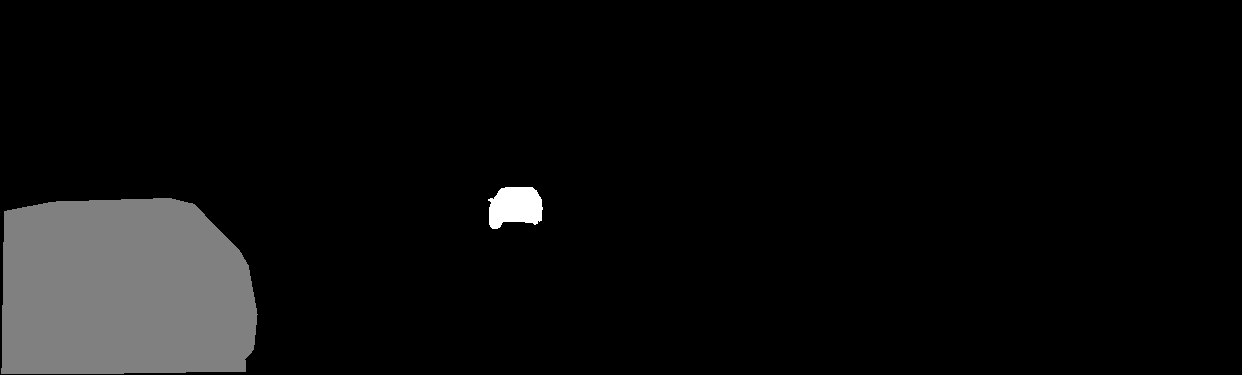}
\includegraphics[width=\MyImgWidth]{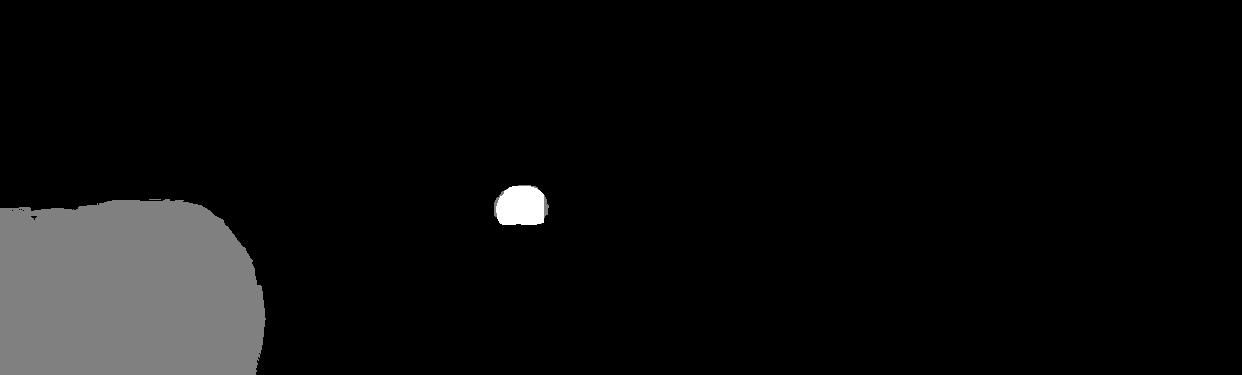}

\includegraphics[width=\MyImgWidth]{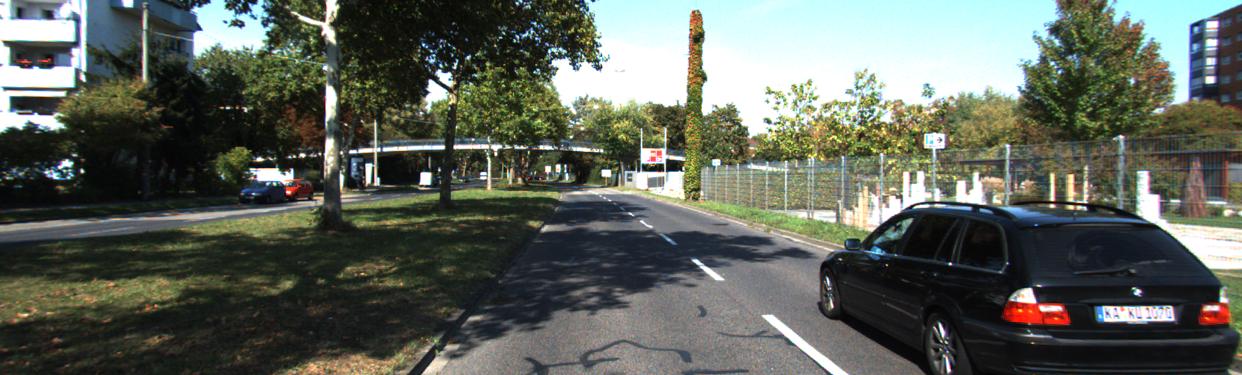}
\includegraphics[width=\MyImgWidth]{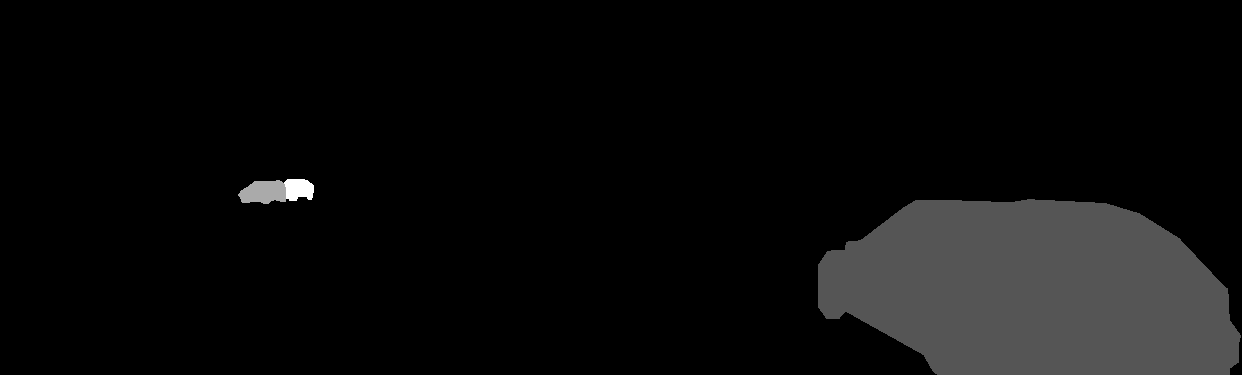}
\includegraphics[width=\MyImgWidth]{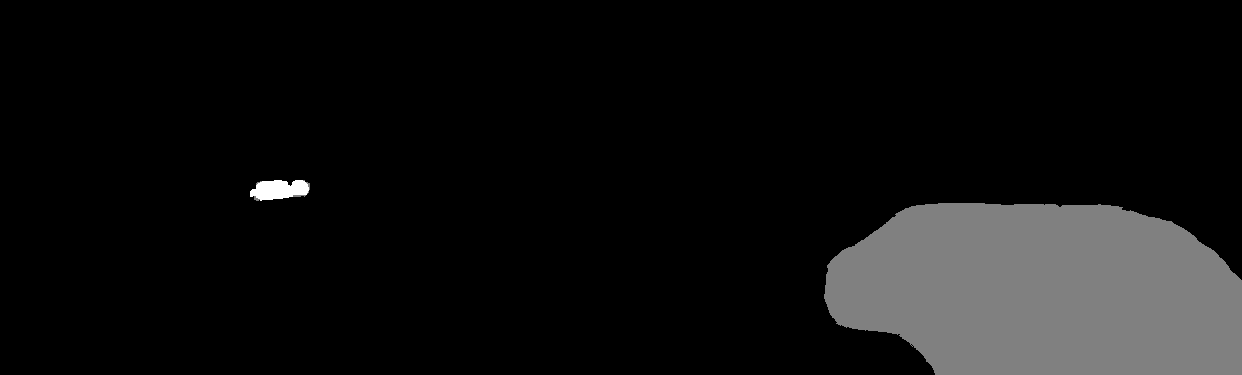}

\includegraphics[width=\MyImgWidth]{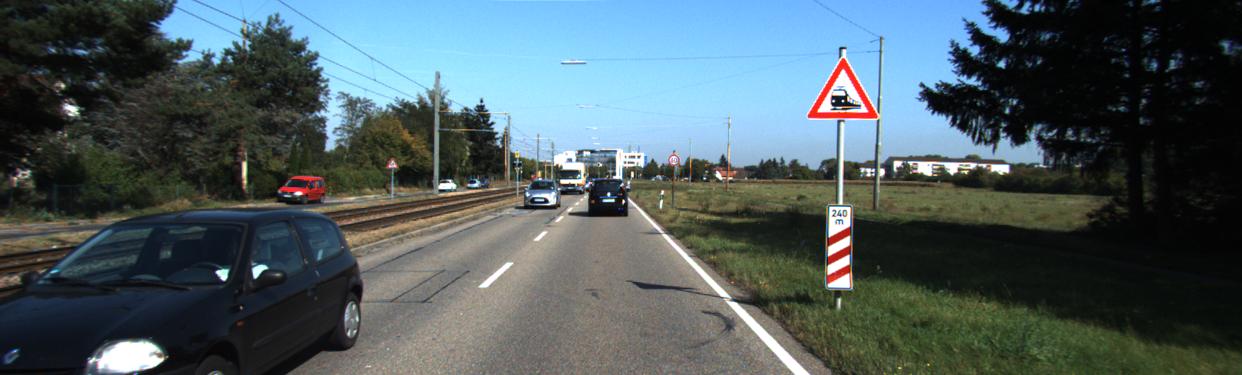}
\includegraphics[width=\MyImgWidth]{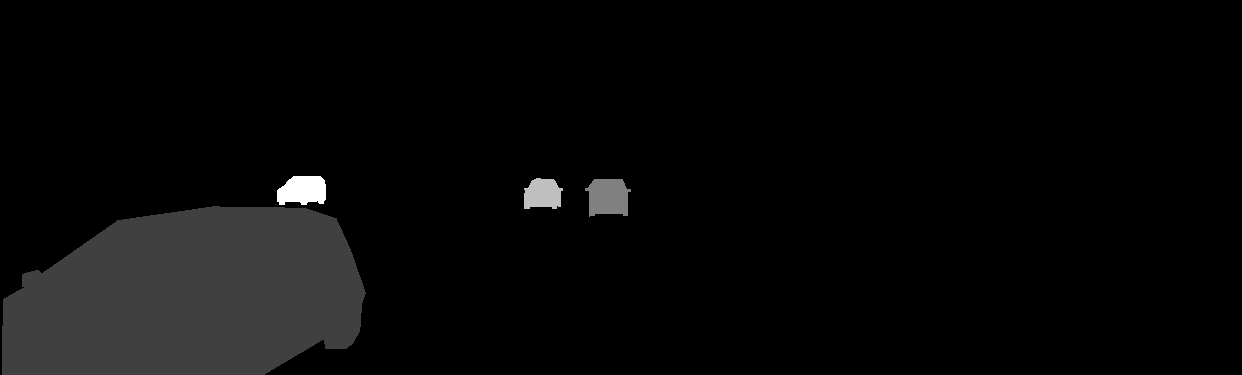}
\includegraphics[width=\MyImgWidth]{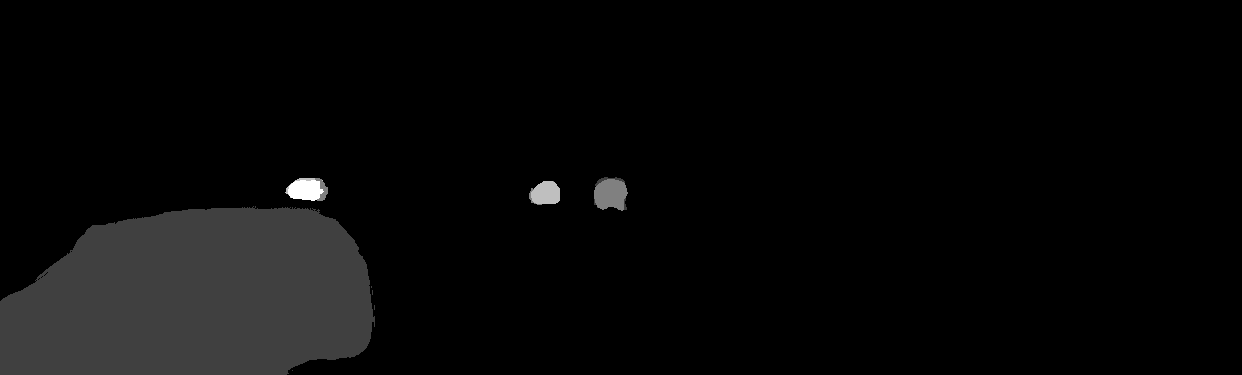}

\includegraphics[width=\MyImgWidth]{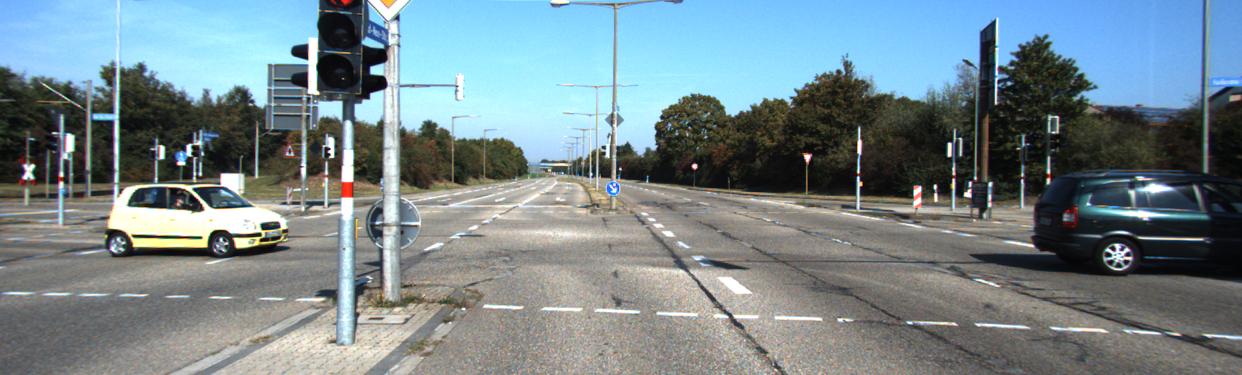}
\includegraphics[width=\MyImgWidth]{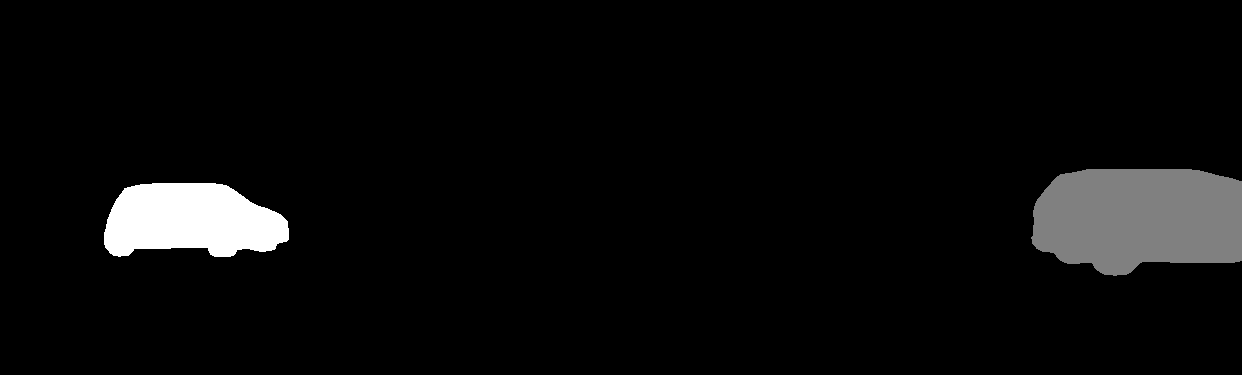}
\includegraphics[width=\MyImgWidth]{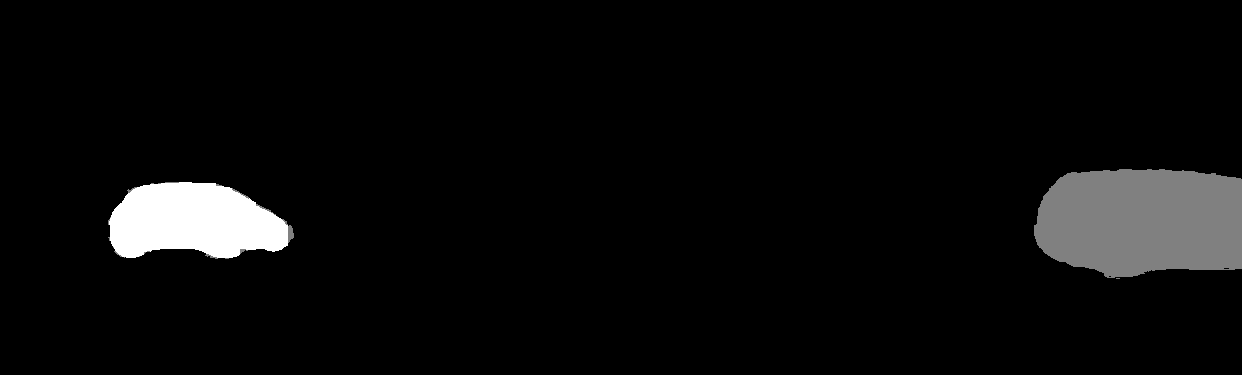}

\includegraphics[width=\MyImgWidth]{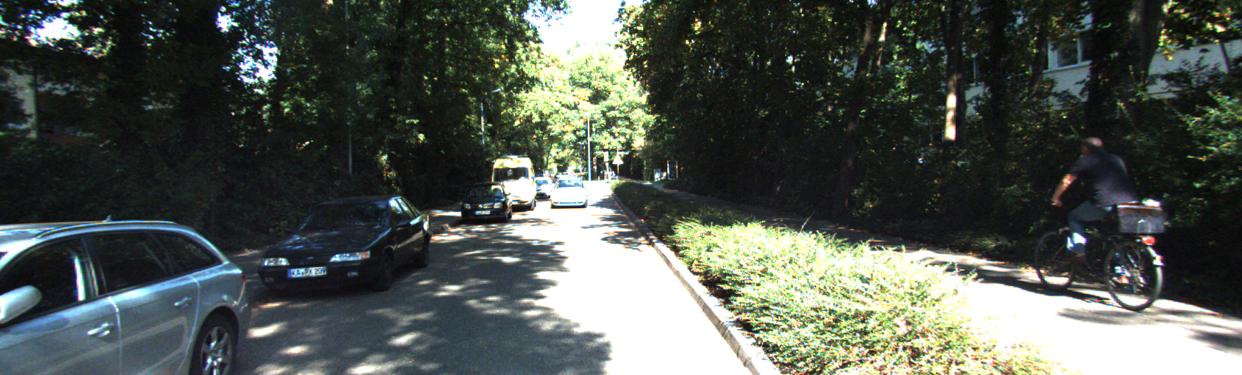}
\includegraphics[width=\MyImgWidth]{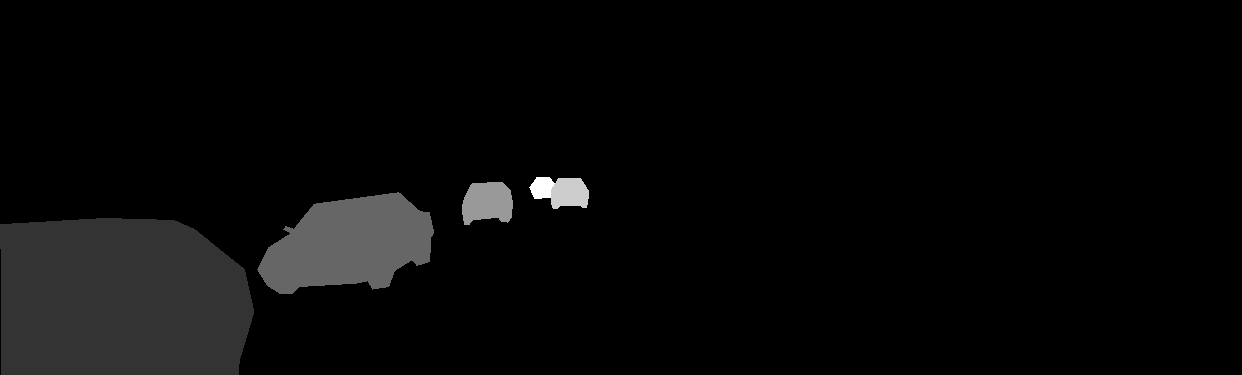}
\includegraphics[width=\MyImgWidth]{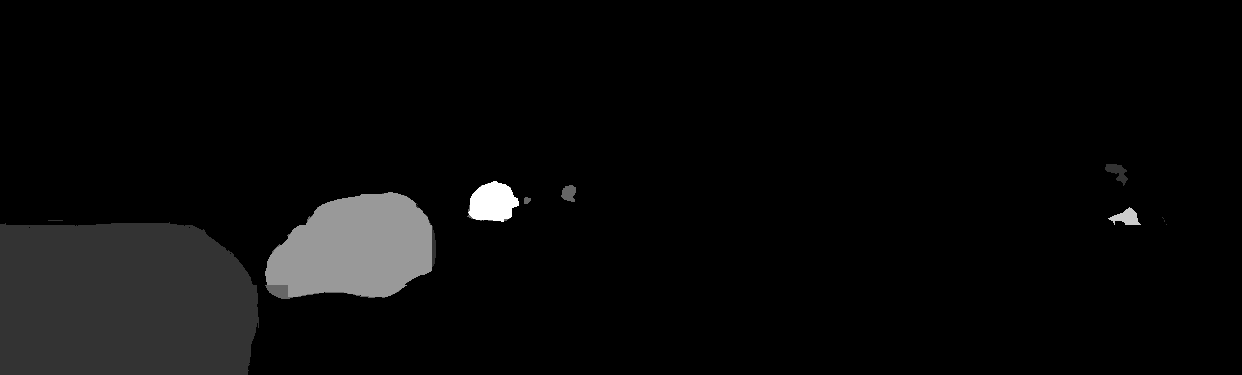}

\includegraphics[width=\MyImgWidth]{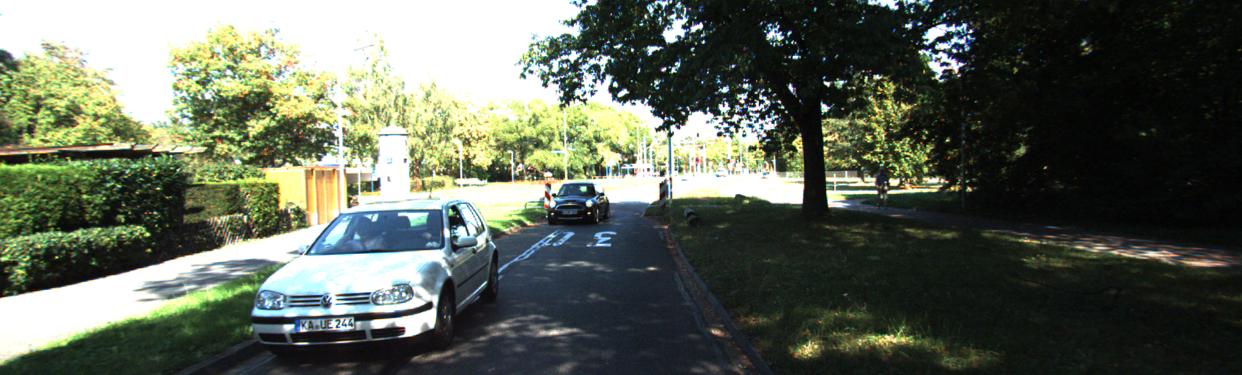}
\includegraphics[width=\MyImgWidth]{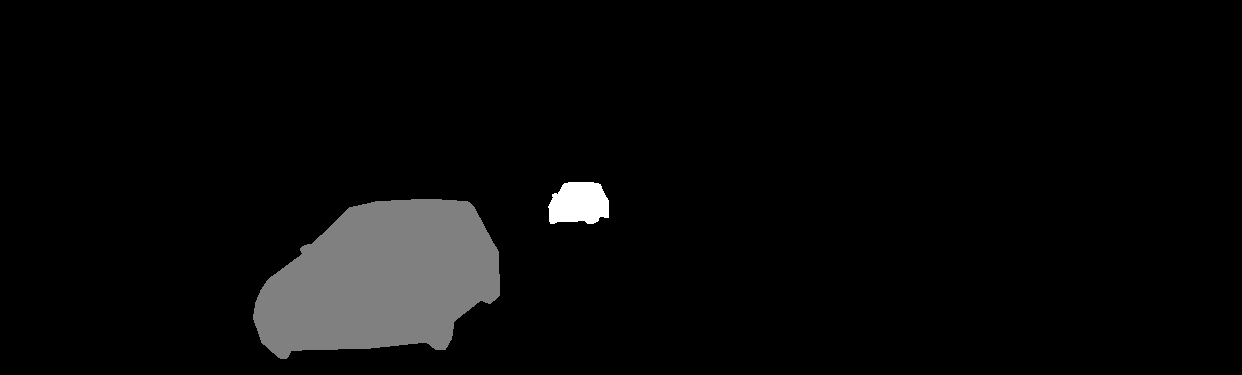}
\includegraphics[width=\MyImgWidth]{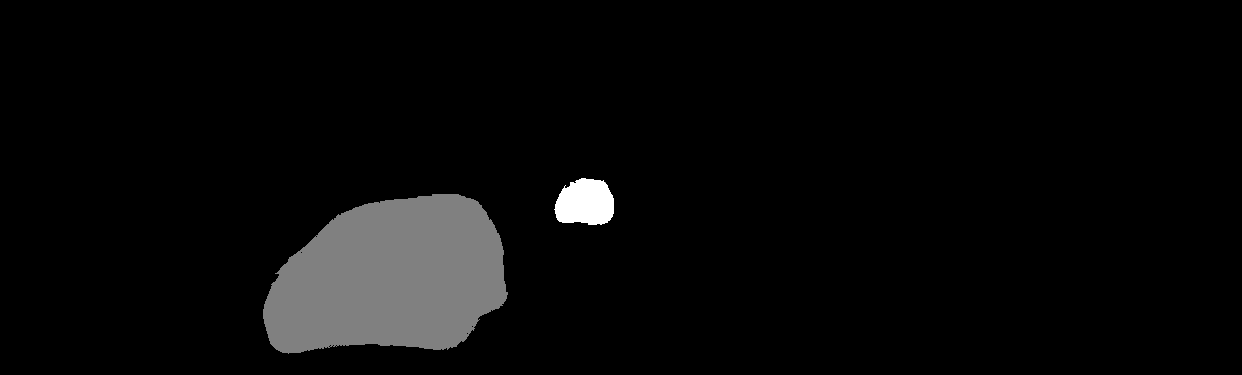}

\includegraphics[width=\MyImgWidth]{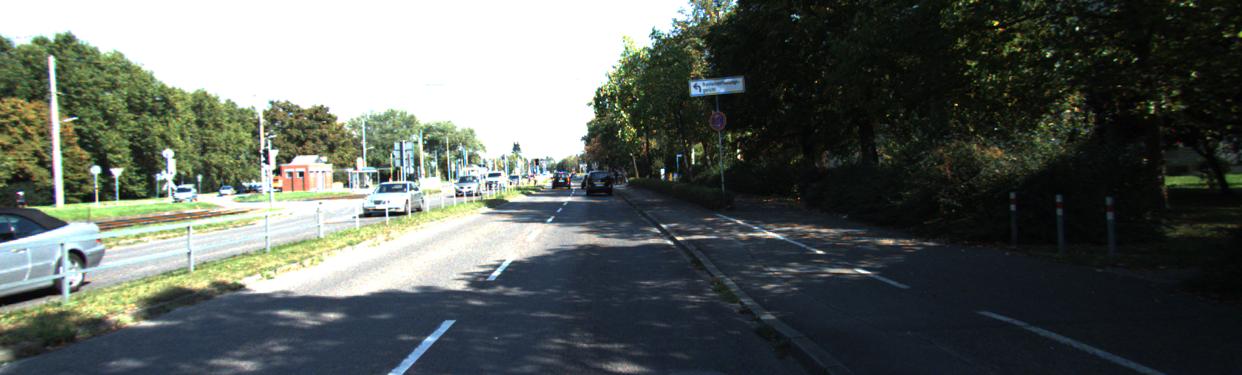}
\includegraphics[width=\MyImgWidth]{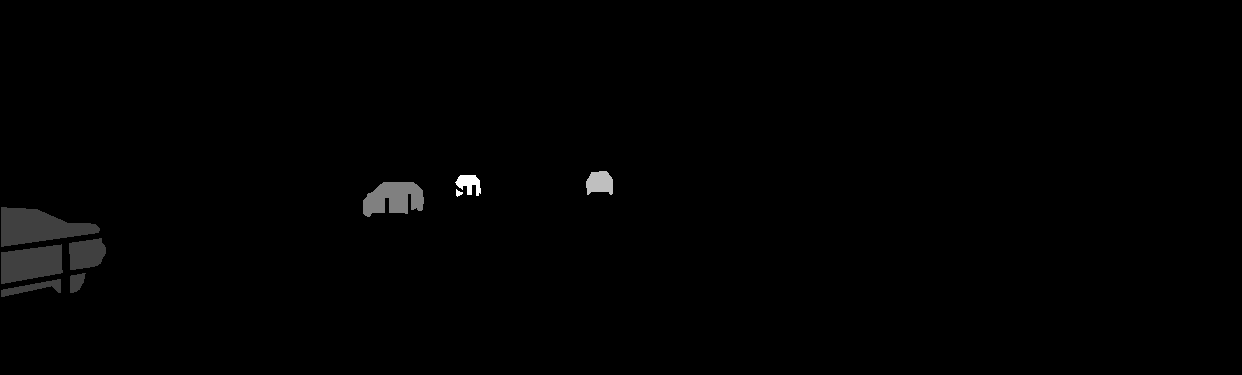}
\includegraphics[width=\MyImgWidth]{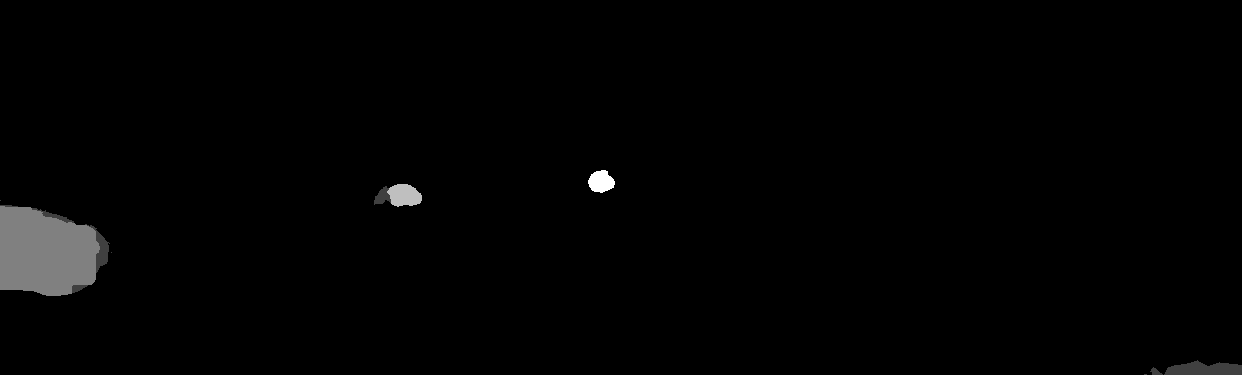}

\includegraphics[width=\MyImgWidth]{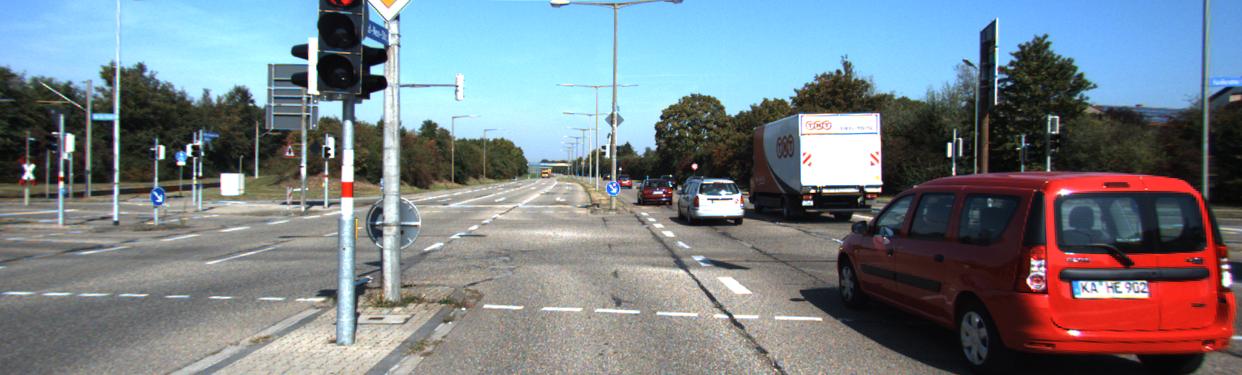}
\includegraphics[width=\MyImgWidth]{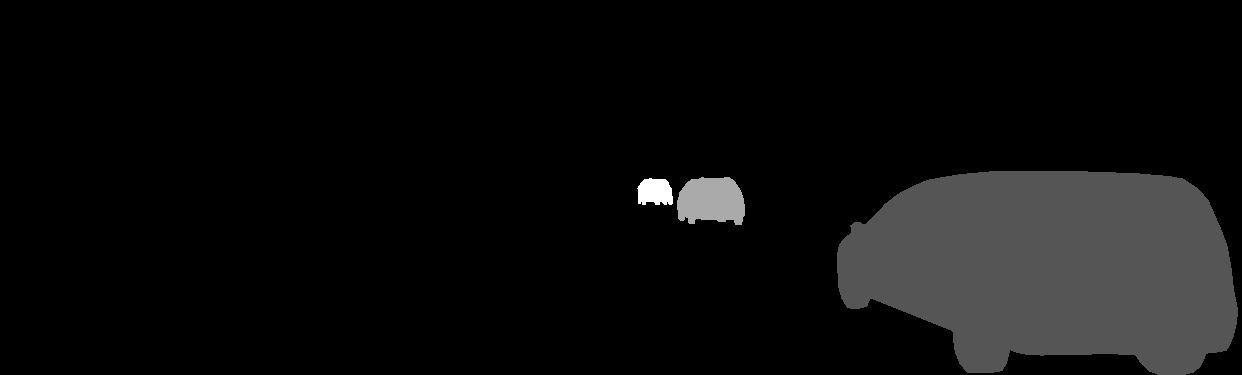}
\includegraphics[width=\MyImgWidth]{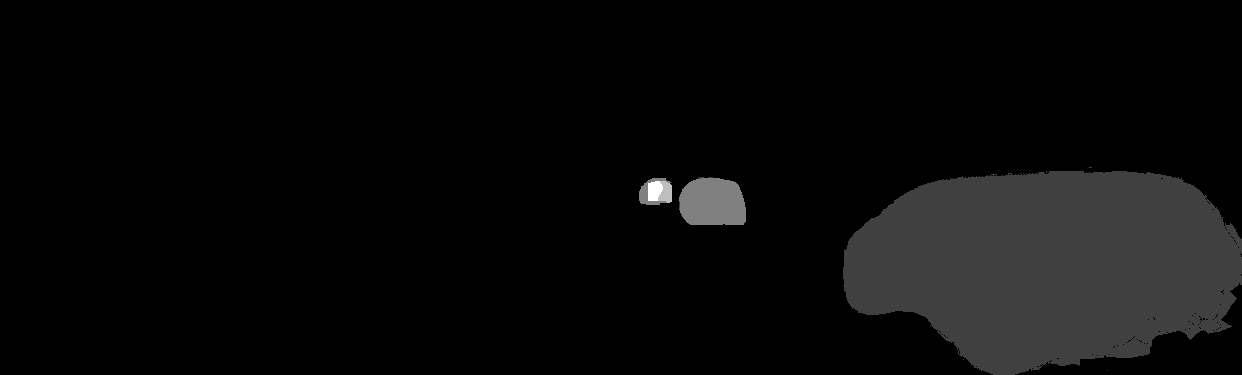}

\includegraphics[width=\MyImgWidth]{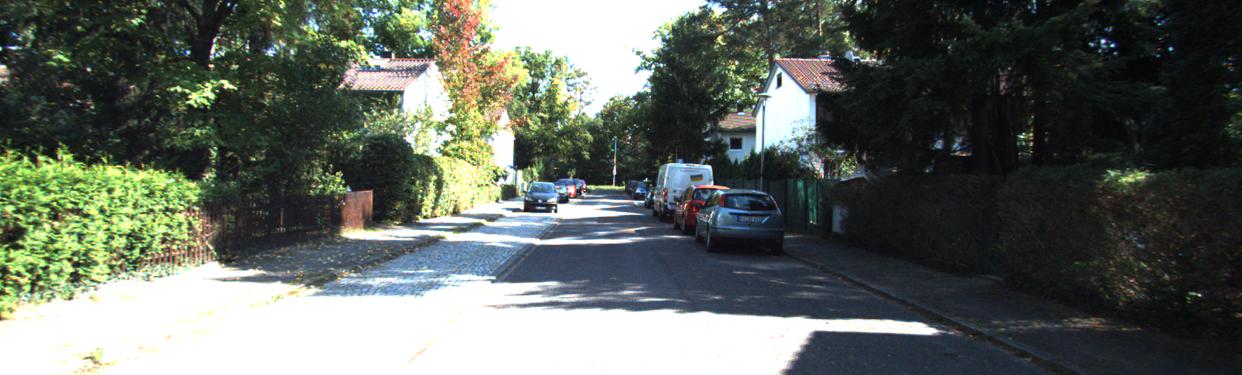}
\includegraphics[width=\MyImgWidth]{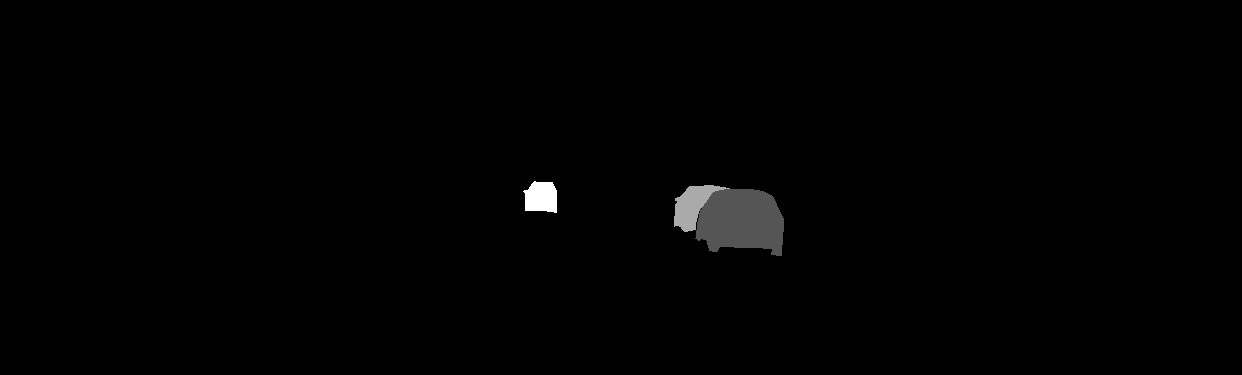}
\includegraphics[width=\MyImgWidth]{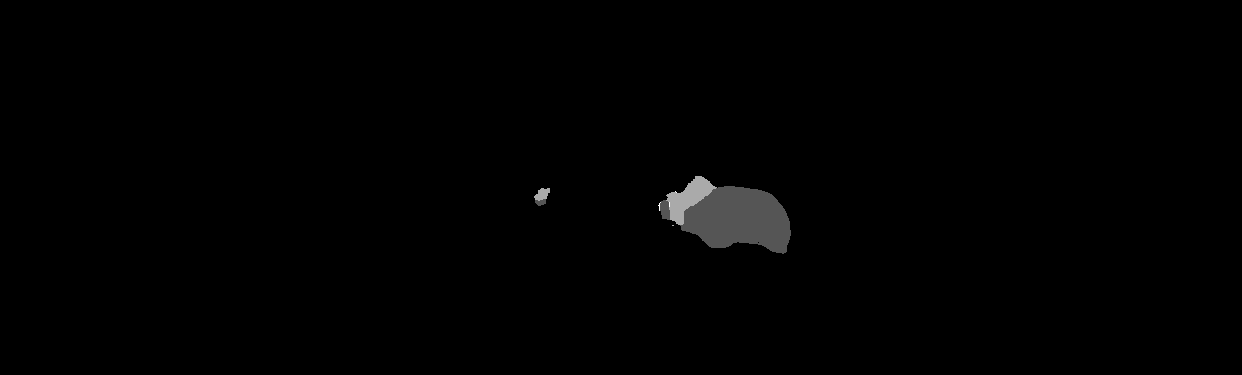}

\includegraphics[width=\MyImgWidth]{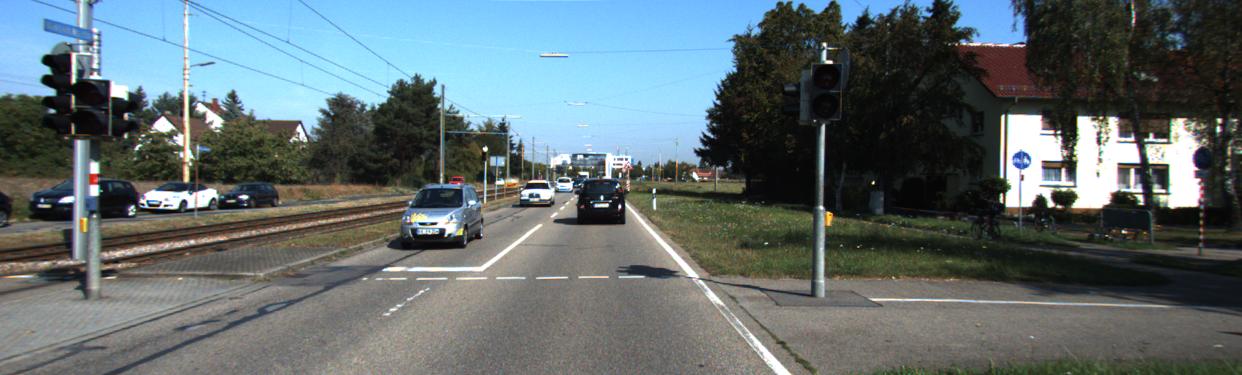}
\includegraphics[width=\MyImgWidth]{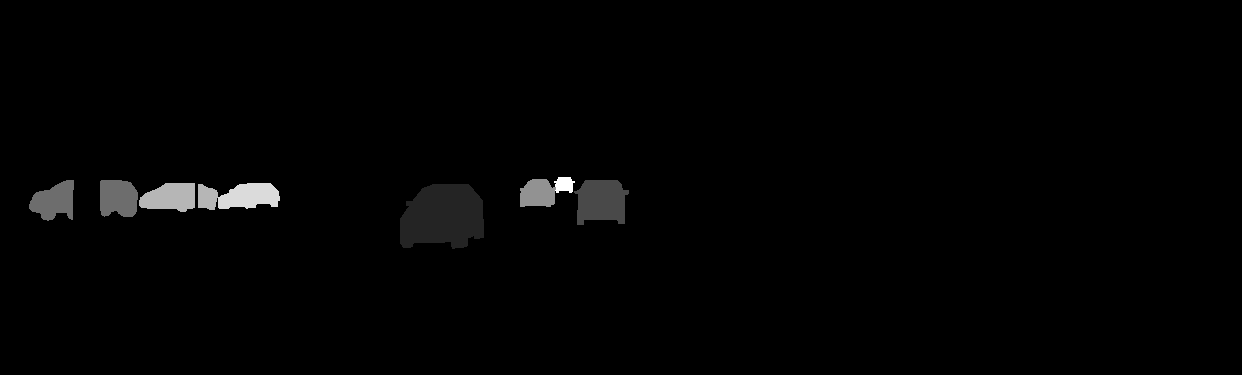}
\includegraphics[width=\MyImgWidth]{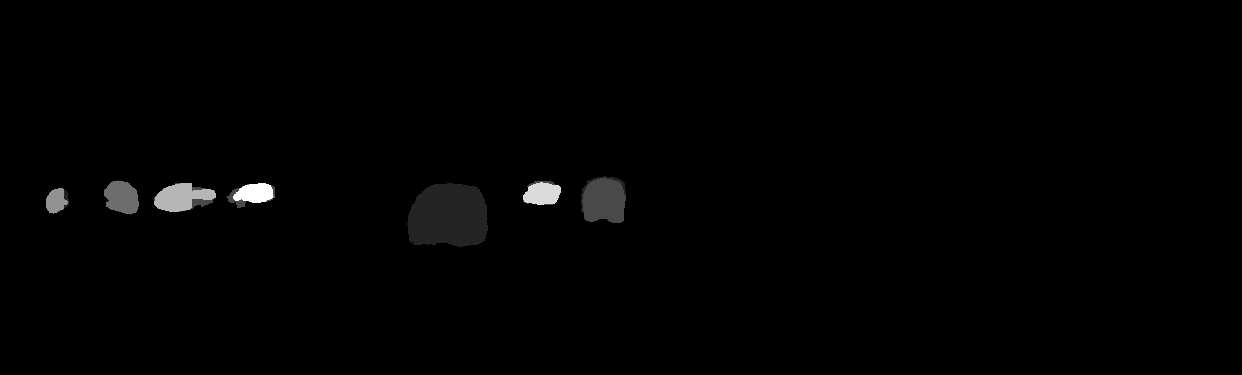}

\includegraphics[width=\MyImgWidth]{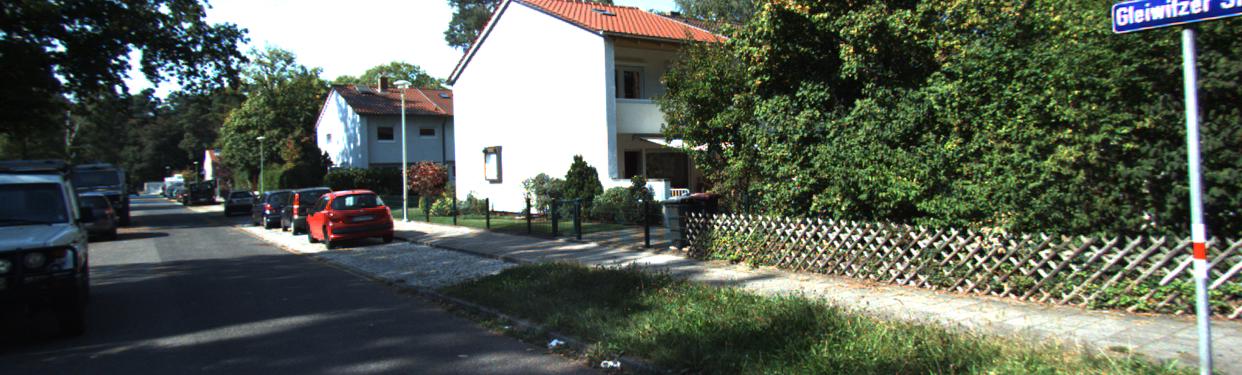}
\includegraphics[width=\MyImgWidth]{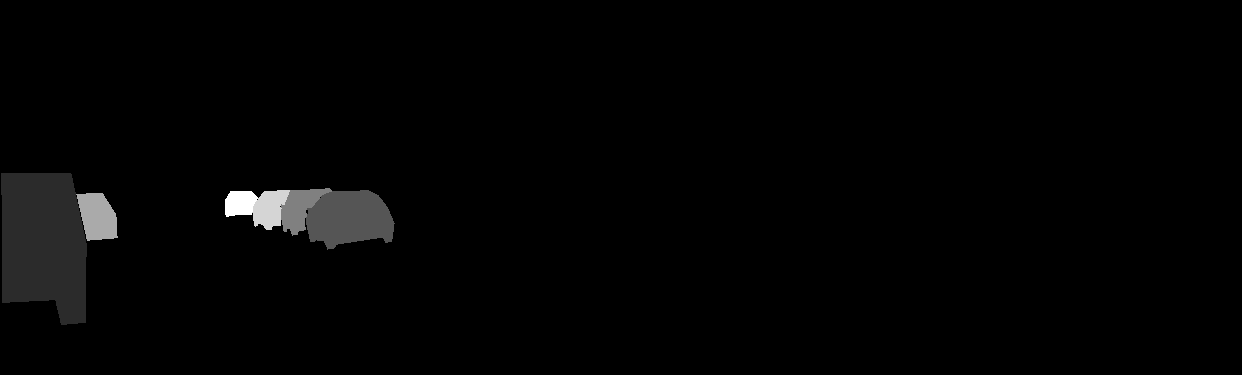}
\includegraphics[width=\MyImgWidth]{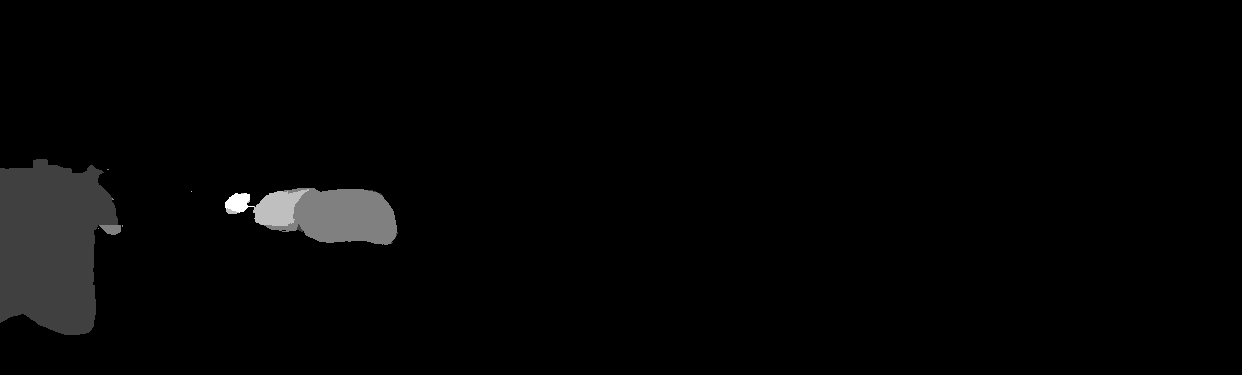}

\includegraphics[width=\MyImgWidth]{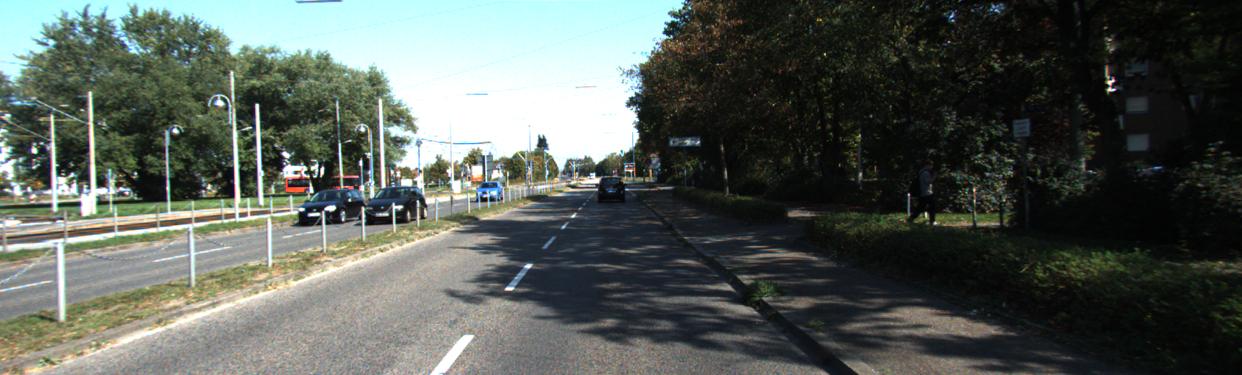}
\includegraphics[width=\MyImgWidth]{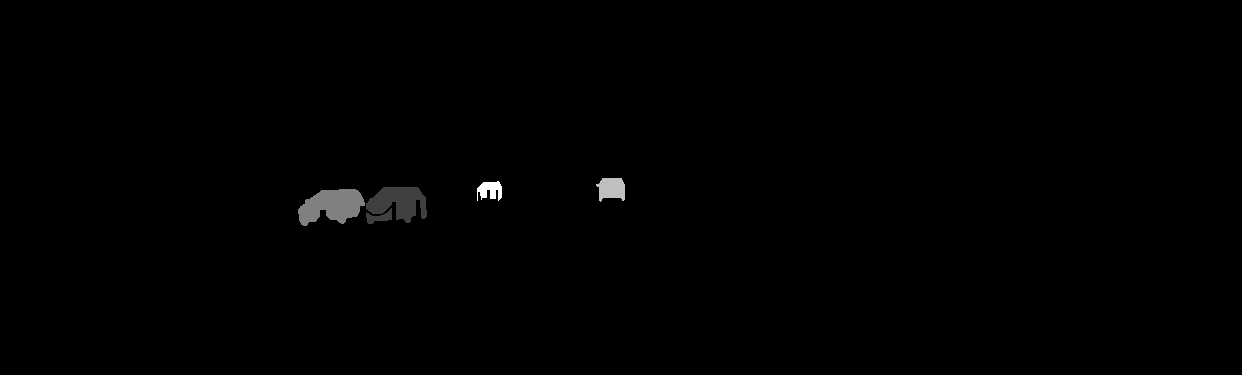}
\includegraphics[width=\MyImgWidth]{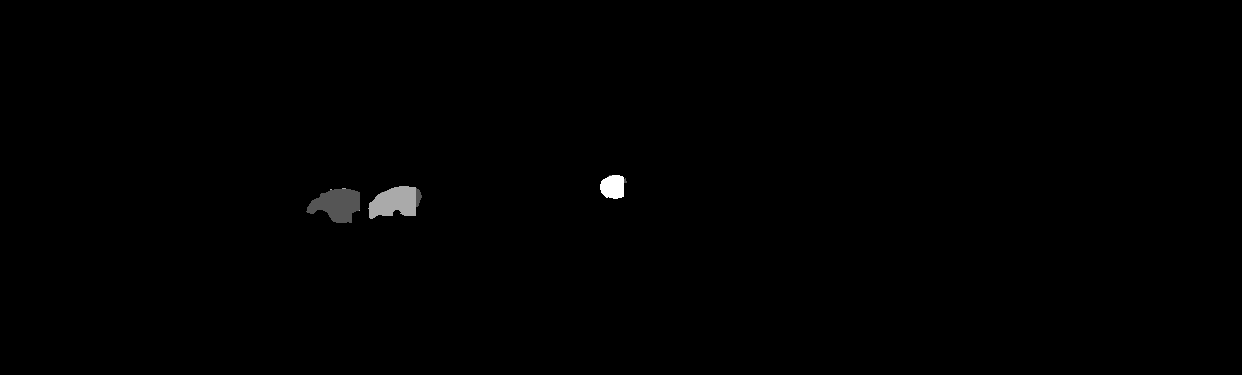}

\caption{Successful prediction compared to the ground truth (middle) given the image (left).}
\label{fig:Success}
\vspace{-0.4cm}
\end{figure*}

\begin{figure*}[t]
\vspace{-0.4cm}
\setlength{\MyImgWidth}{5cm}
\centering

\includegraphics[width=\MyImgWidth]{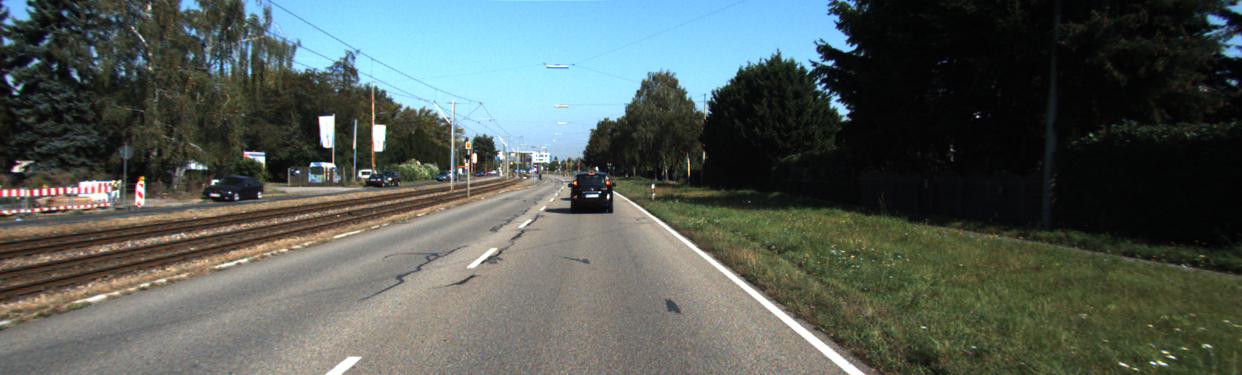}
\includegraphics[width=\MyImgWidth]{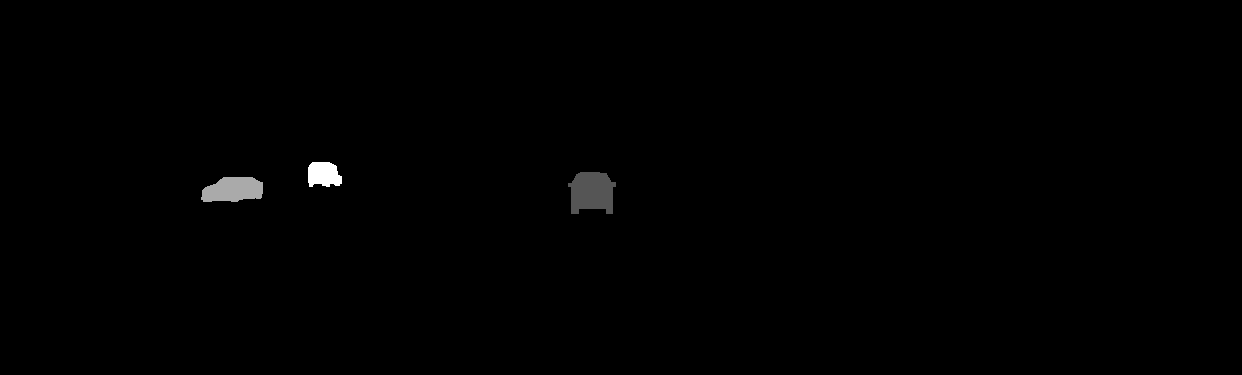}
\includegraphics[width=\MyImgWidth]{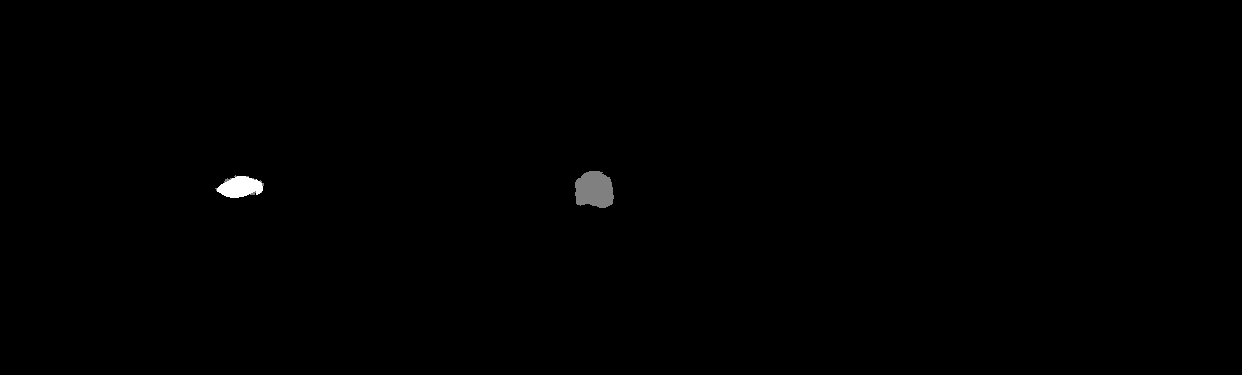}

\includegraphics[width=\MyImgWidth]{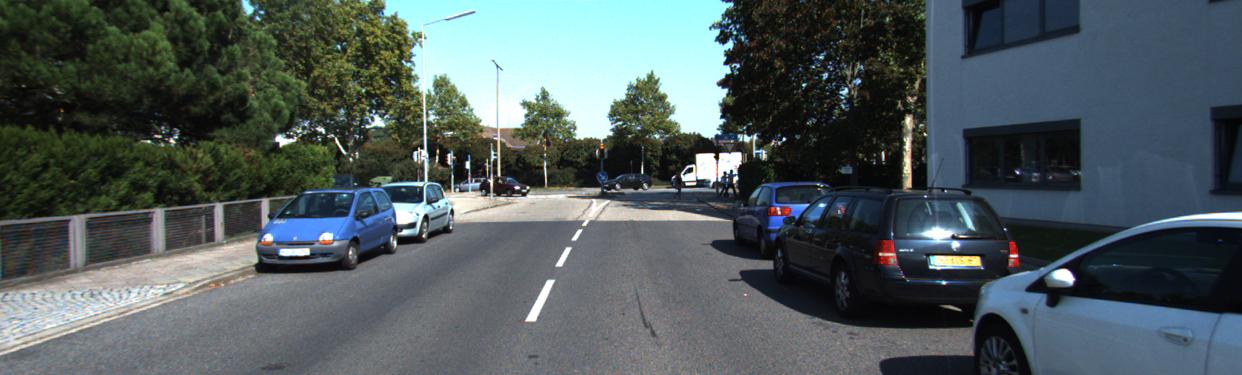}
\includegraphics[width=\MyImgWidth]{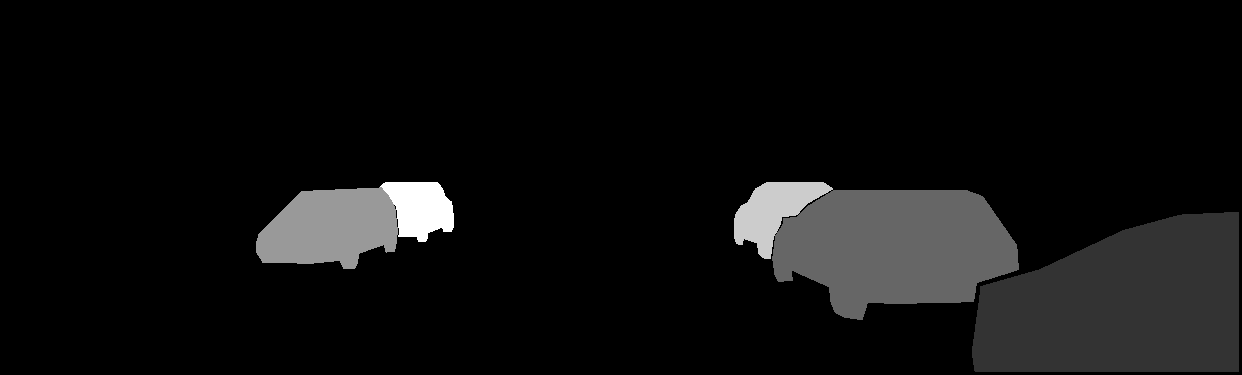}
\includegraphics[width=\MyImgWidth]{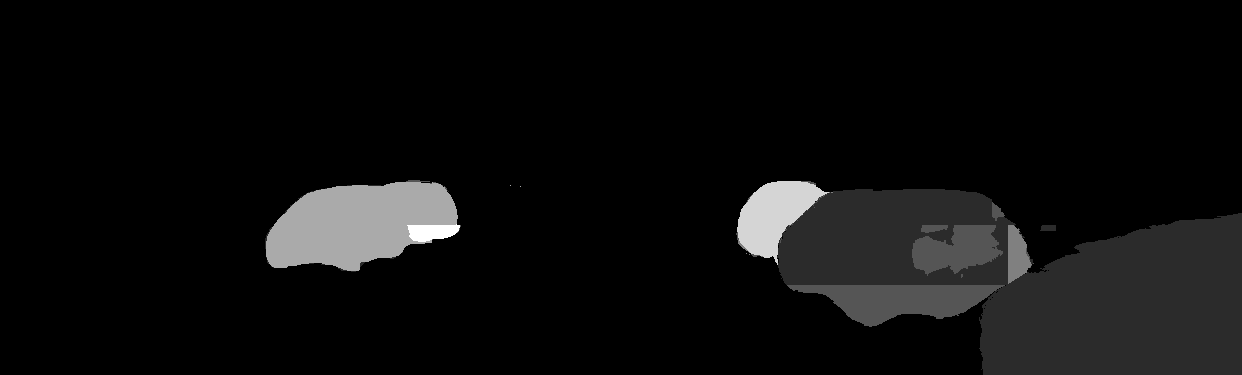}

\includegraphics[width=\MyImgWidth]{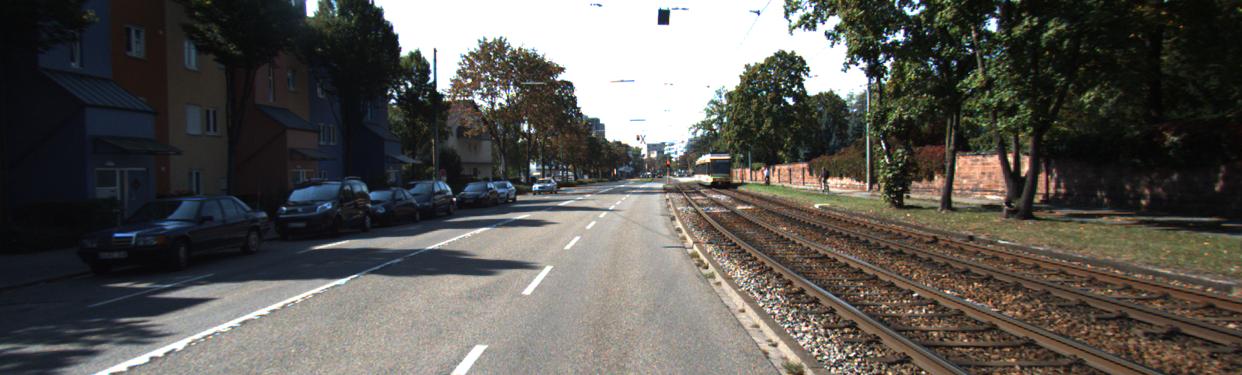}
\includegraphics[width=\MyImgWidth]{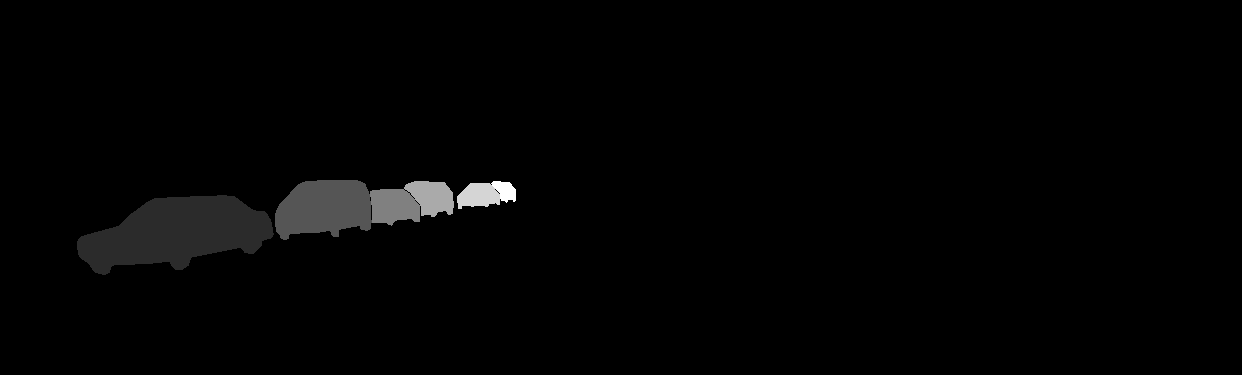}
\includegraphics[width=\MyImgWidth]{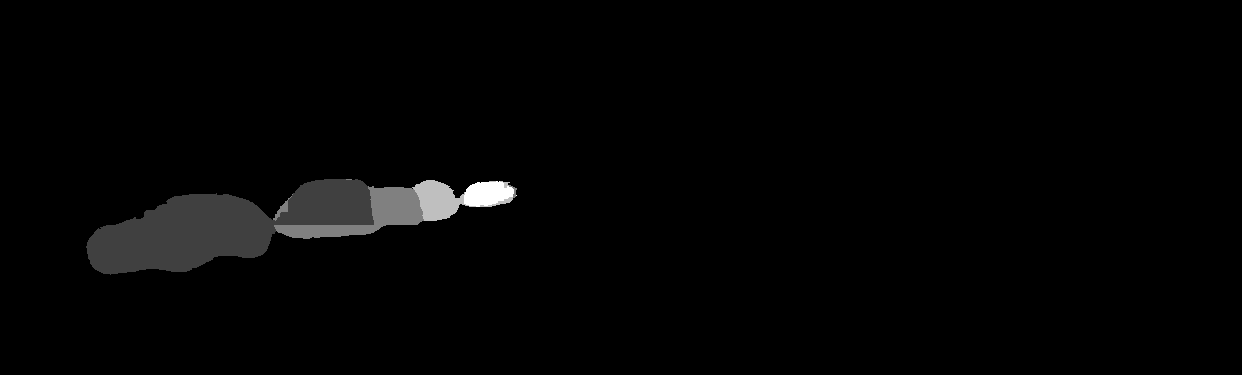}

\includegraphics[width=\MyImgWidth]{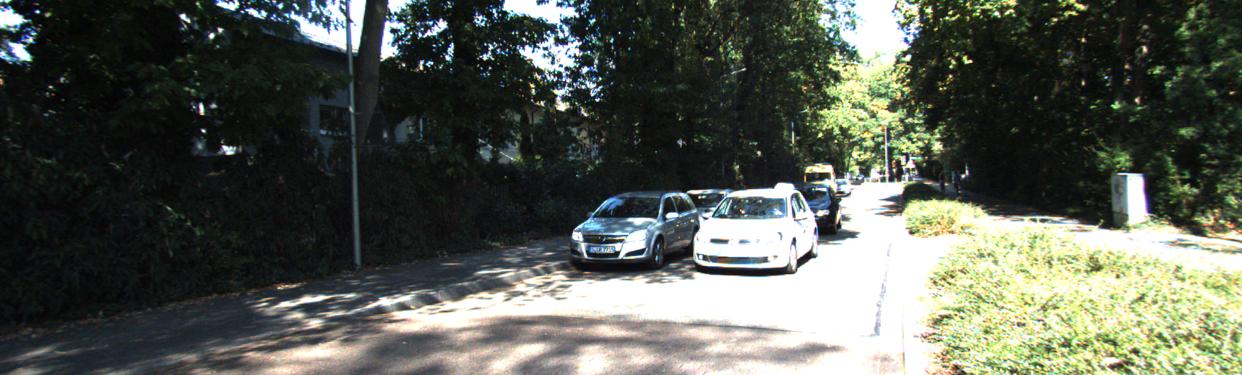}
\includegraphics[width=\MyImgWidth]{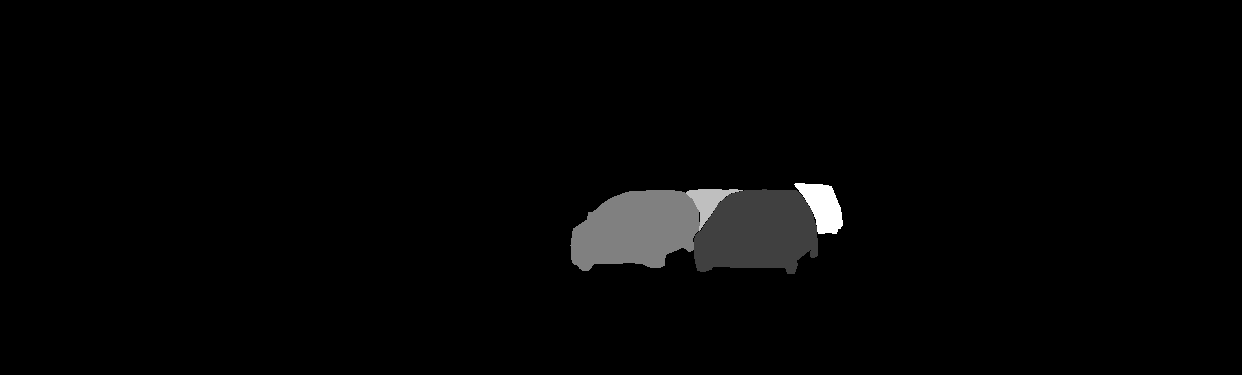}
\includegraphics[width=\MyImgWidth]{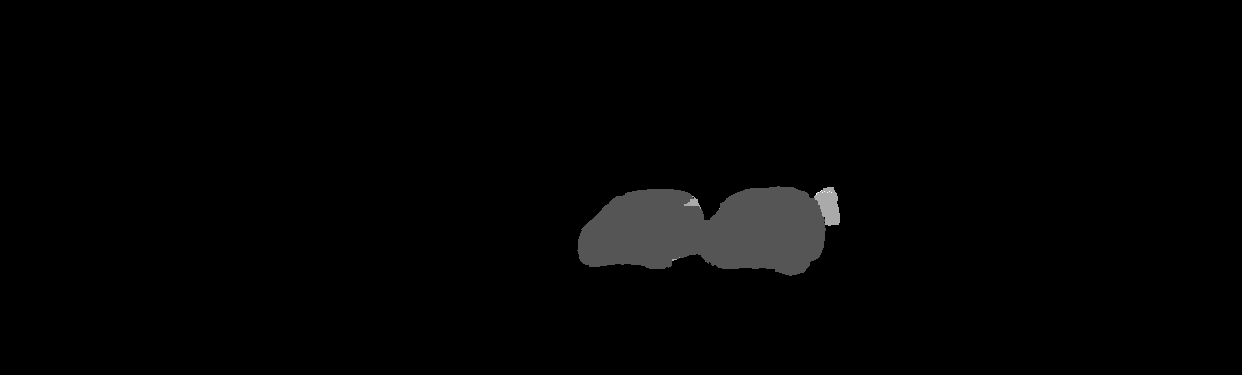}

\caption{Failure cases compared to the ground truth (middle) given the image (left).}
\label{fig:Failure}
\vspace{-0.4cm}
\end{figure*}

\vspace{-0.4cm}
\paragraph{Instance segmentation:} We evaluate instance-level segmentation  using a variety of metrics including mean weighted coverage (MWCov), mean unweighted coverage (MUCov), average precision (AvgPr), average recall (AvgRe), average false positives (AvgFP) and average false negative (AvgFN), as well as object precision (ObjPr) and object recall (ObjRe).
Mean weighted coverage and mean unweighted coverage was introduced by~\cite{HoiemIJCV2011,SilbermanECCV14}. For each ground truth region we find the maximally possible IoU score using the predictions. We obtain the weighted and unweighted coverage by summing the maximum IoU scores for all ground-truth regions, in the weighted case multiplied by the ground-truth region size.
We evaluate the average precision/recall on a pixel-level for every predicted/ground-truth instance and average their results. 
We compute false positives and false negatives as described in the following. If a prediction does not overlap with any ground-truth instance, it is a false positive. The average false positive metric is then the number of false positive object instances within an image averaged across images. Similarly for false negatives where a ground-truth instance has no overlap with any prediction. 
The object precision and recall are evaluated purely on the instance level. For each ground-truth instance we find a prediction with IoU larger than $50\%$, \ie, the number of correctly classified instances, and either we divide this number by the total number of predictions (ObjPr) or we divide it by the total number of ground-truth instances (ObjRe).
We provide the numbers for the test set in \tabref{tab:InstanceMeasuresTest}. We note that again the medium sized patches perform very well. Further, we observe that the baseline again achieves  good precision but very bad recall, while our full MRF approach performs well on all other metrics compared to the baseline. We want to particularly highlight the `MWCov' and `MUCov' metrics which are true instance level evluation where we outperform the baseline significantly. Importantly we note that our post-processing step improves performance significantly by around $2\%$.
In our ablation analysis we observe that unary performance outperforms the full approach when no post-processing is applied. After post-processing the pairwise MRF formulation clearly outperforms the usage of unary potentials only.

\vspace{-0.4cm}
\paragraph{Depth ordering:} We evaluate depth ordering quality using yet another set of metrics.
Given the total number of object instances (\#Ins), we provide the number of correctly predicted instances (\%RcldIns), \ie, this is the fraction of instances for which we are able to find a ground truth instance having a IoU larger than $50\%$.
In addition we indicate the number of possible object instance pairs (\#InsPair) and a metric measuring for how many ground-truth pairs we are able to find predictions for both having an IoU larger than $50\%$ (\%RcldInsPair). Instance pair accuracy (InsPairAcc) measures how many recalled instance pairs are also predicted correctly in the depth ordering. 
We also compute the percentage of correctly classified foreground pixel pairs (\%CorrPxlPairFgr). Since this is oftentimes a very large number of possible pairs, we randomly sample 5\% of the all possible pixel pairs and check their correctness. Note that  5\% of the pairs is still around 30 million pairs per image on average.
We provide the depth ordering evaluation in \tabref{tab:DepthMeasuresTest}.

To assess the quality of depth ordering we compare our approach to three baselines each ordering the predicted instances of~\cite{ohn2015learning} with a different method. The first method orders the detected instances by their lowest pixel location along the vertical axis (\cite{ohn2015learning}+Y). The second baseline orders detected instances by their mean depth value (\cite{ohn2015learning}+Depth). The depth map for each image is computed using the depth-from-single-image method of~\cite{karsch}. The third approach orders the instances by their size (\cite{ohn2015learning}+Size).



We observe our approach to work reasonably well, getting $83.1\%$ of the randomly sampled foreground pixel pairs ordered correctly.
Using an ablation analysis we note that our pairwise MRF formulation improves upon the raw CNN output and its converted unaries only after post-processing. Indeed, without post-processing, the pairwise connections harm performance. 

\vspace{-0.4cm}
\paragraph{Qualitative results: }
We provide successful test set predictions in \figref{fig:Success} and illustrate failure modes in \figref{fig:Failure}. Our method performs well on scenes where cars are easily separable as, \eg, illustrated in the top rows of \figref{fig:Success}. This is to a large extent due to the connected component algorithm which is successfully able to disambiguate the instances. On the other hand we observe challenges for tiny cars, \eg, missed by the CNN prediction, and instances that are merged by the connected component algorithm. Illustrations for the former case are given in row 1, while merged instances are illustrated in row 2 through 4 of \figref{fig:Failure}.

\section{Conclusion}

In this paper we have presented an approach that exploit a convolutional neural network as well as a Markov random field to produce accurate instance level segmentations and depth orderings from a single monocular image. We have demonstrated the effectiveness of our approach on the challenging KITTI benchmark and show that we can learn to segment from weak annotations in the form of 3D bounding boxes. 
In the future we plan to apply our approach to indoor scenes, which might have  higher degree of clutter.

\noindent{\bf Acknowledgments:} We thank NVIDIA Corporation for the donation of GPUs used in this research. This work was partially funded by ONR-N00014-14-1-0232.

{\small
\bibliographystyle{ieee}
\bibliography{egbib}
}

\end{document}